%% file: TNNLS-2012-P-0824.tex
\documentclass[journal]{IEEEtran}
\usepackage{flushend}
\usepackage{cite}
\usepackage{array}
\usepackage{subfigure}
\usepackage{graphicx}
\usepackage{amsmath}
\usepackage{amssymb}
\usepackage{mathptmx}
\usepackage{epsfig}
\usepackage{color}
\usepackage[normalem]{ulem}
\usepackage{wasysym}
\usepackage{fancyhdr}
\usepackage{multirow}

\begin{document}
%
\title{Fast Neuromimetic Object Recognition using FPGA Outperforms GPU Implementations}


\author{ {Garrick Orchard, Jacob G. Martin, R. Jacob Vogelstein, and Ralph Etienne-Cummings, \textit{Fellow, IEEE}}
\thanks{Manuscript received September 2, 2011, revised January 10, 2012, revised April 10, 2012, revised June 10, 2012, revised Sept 15, 2012, revised Nov 15, 2012, revised March 5, 2013, accepted March 15, 2013.
The views and conclusions contained in this document are those of the authors and should not be interpreted as representing the official policies, either expressly or implied, of the Defense Advanced Research Projects Agency of the U.S. Government.}
\thanks{Garrick Orchard, R. Jacob Vogelstein, and Ralph Etienne-Cummings are with the Department of Electrical and Computer Engineering at The Johns Hopkins University, Baltimore, Maryland 21218 USA (Email: gorchard@jhu.edu, jacob.vogelstein@jhuapl.edu, retienne@jhu.edu)}
\thanks{Garrick Orchard is also with the Singapore Institute of Neurotechnology at the National University of Singapore in Singapore, 117456, Singapore (Email:garrickorchard@nus.edu.sg)}
\thanks{Jacob G. Martin and R. Jacob Vogelstein are with the The Johns Hopkins University Applied Physics Laboratory in Laurel, Maryland 20723 USA (Email: jacob.martin@jhuapl.edu, jacob.vogelstein@jhuapl.edu)}
}

\maketitle

\begin{abstract}

Recognition of objects in still images has traditionally been regarded as a difficult computational problem.  Although modern automated methods for visual object recognition have achieved steadily increasing recognition accuracy, even the most advanced computational vision approaches are unable to obtain performance equal to that of humans.  This has led to the creation of many biologically-inspired models of visual object recognition, among them the HMAX model. HMAX is traditionally known to achieve high accuracy in visual object recognition tasks at the expense of significant computational complexity.  Increasing complexity, in turn, increases computation time, reducing the number of images that can be processed per unit time.  In this paper we describe how the computationally intensive, biologically inspired HMAX model for visual object recognition can be modified for implementation on a commercial Field Programmable Gate Array, specifically the Xilinx Virtex 6 ML605 evaluation board with XC6VLX240T FPGA. We show that with minor modifications to the traditional HMAX model we can perform recognition on images of size 128$\times$128 pixels at a rate of 190 images per second with a less than 1\% loss in recognition accuracy in both binary and multi-class visual object recognition tasks.

\end{abstract}


%
\IEEEpeerreviewmaketitle

\section{Introduction}
Object recognition has received a lot of attention in recent years and is an important step towards building machines which can understand and interact meaningfully with their environment. In this context, both a high recognition accuracy and a short recognition time are desirable. By shortening recognition time even further, we foresee applications that include rapidly searching and categorizing images on the internet based on features extracted from their pixel content on the fly.  Many currently available image search and characterization platforms rely on image metadata and watermarks rather than the images' actual pixel values, while those platforms which do make use of actual pixel values typically rely on previously extracted image features rather than creating and extracting new features on the fly.

The challenge of consistently recognizing an object is complicated by the fact that the appearance of the object can vary significantly depending on its location, orientation, and scale within an image. Reliable object recognition must therefore be invariant to translation, scale, and orientation. Some methods of object recognition incorporate these invariances, such as the Scale Invariant Feature Transformation (SIFT) \cite{SIFT} or Speeded Up Robust Features (SURF) \cite{SURF}. These models achieve good recognition rates, but still fall far short of the recognition rates achieved by humans. There is evidence suggesting that after viewing an object for the first time, a biological system is capable of recognizing that object again at a novel position and scale \cite{MonkeyViewPaperclip}. The object can also be recognized if it is slightly rotated, but the recognition accuracy decreases when the object is rotated too far from a familiar view \cite{MonkeyViewPaperclip}. A biologically inspired model which shares this property of scale and translation invariance, but also achieves only limited rotation invariance is the Hierarchical Model and X (HMAX) \cite{HMAXOriginal}  in which the `X' represents a non linearity.

Jarrett~\emph{et al.} \cite{BestArchitectureForRecognition} investigated which architecture is best for object recognition. They found that non-linearities are the most important feature in such models. Their results show that rectification and local normalization significantly improve recognition accuracy. Their results also indicate that a multistage method of feature extraction outperforms single stage feature extraction. The HMAX model is a multistage model which mixes Gabor filters in the first stage with learned filters in the second. HMAX is intended to model the first 100-200ms of object recognition due to purely feed-forward mechanisms in the ventral visual pathway \cite{HMAXOriginal}. HMAX is biologically inspired and incorporates rectification and local-normalization non-linearities, both of which were later recommended by Jarret~\emph{et al.} \cite{BestArchitectureForRecognition} as important properties for object recognition models.

In this paper, we focus specifically on the version of HMAX described in \cite{HMAX2007}. The recognition accuracy of HMAX is well below that of the biological counterparts it attempts to mimic for real world tasks because it only mimics the first stages of the feed-forward pathways. However, HMAX performs comparably to its biological counterparts on rapid characterization tasks in which a stimulus is presented long enough for feed forward recognition to take place, but short enough to prevent top down feedback from having an effect \cite{FeedforwardRapidCategorization, AnimalNoAnimal}. HMAX provides a valuable step towards achieving higher recognition accuracy and better understanding the operation of the ventral stream in visual cortex. Biological processing systems (networks of neurons) are inherently distributed and massively parallel. If we intend to achieve comparable recognition rates by mimicking biological processing, then we too should use distributed and massively parallel hardware which is suited to the task.

Originally, object recognition models were typically run on sequential processors (CPUs), for which Mutch and Lowe developed the Feature Hierarchy Library (FHLib) tool in 2006 \cite{MUTCH06} for implementing hierarchical models such as HMAX. CPUs require little effort to program and offer great flexibility, allowing them to be used for a large variety of tasks, but the sequential nature of their processing makes them ill suited to an application such as HMAX. Modern CPUs are capable of impressive performance and allow some parallel processing, but depending on the nature of the algorithm to be implemented, it can be very difficult, if not impossible, to fully utilize the theoretical computational capacity of such devices. In 2008 Chikkerur \cite{CUDA2008} reported a multithreaded CPU implementation of HMAX, showing that the increased parallelism outperformed previous CPU implementations.

{GPUs allow even more parallel processing paths, but writing code for GPUs requires a larger effort than for CPUs}. GPUs also offer greater control of data flow and storage during computation, which allows programmers to make greater use of the theoretical computational capacity. In the same paper as his multithreaded CPU implementation \cite{CUDA2008}, Chikkerur presented a GPU implementation of HMAX with even more parallel processing paths, which outperformed the multithreaded CPU implementation by 3$\times$-10$\times$ depending on input image size. Soon GPU technologies were being used extensively for HMAX and in 2010 Mutch and Lowe released the Cortical Network Simulator (CNS) \cite{GPUhmaxCNS} which uses a GPU for processing and can speedup the HMAX model by 97$\times$ compared to the FHLib software it was intended to replace. Later in 2010, Sedding~\emph{et al.} \cite{CudaTimes2010} presented another GPU implementation of HMAX which is claimed to outperform the CNS implementation in both accuracy and speed. There are also many other examples in the literature of the application of GPU processing to object recognition \cite{GPUobjectRecognition,GPUsift,GPUcudaRecognition,GPUrealTimerecognition}.

Application Specific Integrated Circuits (ASICs) offer an even greater level of control than GPUs through intentional design of the hardware to suit the task at hand, but once fabricated, an ASIC is typically ill suited to other applications. {Furthermore, ASICs require a large design effort, a long time to implement (while waiting for fabrication), and come at high cost, which excludes them from use in many cases.} Nevertheless, high performance still makes ASICs an attractive option for some tasks. An example of such work is the object recognition processor developed by Kim~\emph{et al.} \cite{ASICKaist} which can recognize up to 10 objects at a rate of 60fps at an image size of 640$\times$480 pixels.

Field Programmable Gate Arrays (FPGAs) fall in the space between GPUs and ASICs in terms of time to implementation and level of control. FPGA hardware (fabric) is designed to be highly reconfigurable, thereby giving more control than with GPUs, but the hardware is already fabricated, thereby eliminating the time for fabrication which plagues ASICs. FPGAs also offer an advantage over GPUs in that they can operate in a standalone manner and interface directly with external sensors. A disadvantage of FPGAs is that their use often requires knowledge of a hardware descriptor language (such as Verilog or VHDL) which can be difficult to learn.

In an attempt to make FPGAs more accessible and user friendly, Impulse Accelerated Technologies Inc. \cite{Impulse} has developed a C-to-FPGA compiler to make FPGA acceleration more accessible to those not familiar with hardware design languages. A review of this and other C-to-FPGA approaches can be found in \cite{CtoFPGAsurvey}. The E-lab at Yale is also working on easing the transition to FPGA with the development of ``NeuFlow" \cite{NeuFlowISCAS2010}, an FPGA based system which can be programmed using the easier to learn Lua \cite{Lua} scripting language. This approach significantly reduces time to implementation, but does not necessarily allow the user to fully exploit the performance capabilities of the FPGA. Despite being a valuable tool, the NeuFlow architecture is not well suited to implementing large filters (the original HMAX model requires filters up to 37$\times$37 pixels in size). Other architectures for implementing HMAX on FPGA, developed in parallel with the work in this paper, have been recently published \cite{PENN1, PENN2, PENN3, PENN4, PENN5, PENN6}. These implementations also show considerable speedup over GPU and CPU implementations. Most interesting of these works is a paper from Kestur \emph{et al.} \cite{PENN2} which operates on higher resolution images (2352$\times$1724 pixels), but uses a saliency algorithm to identify regions of interest, thereby obtaining further speedup by circumventing the need for an exhaustive search. Further discussion and comparison with these works can be found in the discussions (Section~\ref{section:Discussion}).

Despite the difficulties of learning hardware design languages, many other vision algorithms have also been implemented in FPGA, including the Lucas-Kanade~\cite{Lucas_Kanade_1981} optical flow algorithm~\cite{FPGAopticalFlow}, SIFT~\cite{SIFTFPGA1, SIFTFPGA2}, SURF~\cite{SURFFPGA} spatiotemporal energy models for tracking~\cite{SpatioTemporalTracking} and segmentation~\cite{SpatioTemporalFPGASegmentation} as well as bioinspired models of gaze and vergence control~\cite{DepthPerceptionVergenceBert}. There are also many examples of Neural Networks (NNs) implemented in FPGA, including multilayer perceptrons \cite{FixedVSfloating}, Boltzmann machines \cite{Boltzmann}, and spiking NNs \cite{SpikingFPGA}.

In work on multilayer perceptrons, Savich~\emph{et al.} \cite{FixedVSfloating} compared the use of fixed point and floating point representations for FPGA implementation and found that fixed point representation used less physical resources, fewer clock cycles, and allowed a higher clock speed than floating point representation while achieving similar precision and functionality. In this work fixed point representation is used throughout.

Himavathi~\emph{et al.} \cite{MultiLayerMultiPlex} described a Neural Network implementation in FPGA which multiplexed resources for computation in different layers, to reduce the total resources required at the expense of computation time. The ultimate aim was to use resources more effectively. In HMAX cells differ by layer, so instead resources are multiplexed for different cells within the same layer. The ultimate aim is similar, to use resources as effectively as possible, thereby achieving maximum throughput with the available resources.

{ The computation performed by the first four layers of HMAX is task independent, allowing us to easily estimate required computation and allocate resources accordingly. The classifier, which follows the fourth HMAX layer, differs depending on the task (binary or multi-class), and in the case of multi-class, the required computation is further dependent on the number of classes (see Section~\ref{section:scalabilityHardware}). To simplify implementation and maintain flexibility of the system, we implement the classification stage in the loop on a host PC. We show through testing in Section~\ref{section:testing} that implementing the classifier in the loop on a host PC does not affect the system throughput. Implementing a classifier in FPGA is nevertheless possible, as is evidenced by numerous examples of FPGA classifier implementations in the literature, including Gaussian Mixture Models (GMMs) \cite{classGMM}, NNs \cite{classNN1, classNN2}, Naive Bayes \cite{classNaiveBayes}, K-Nearest Neighbour (KNN) \cite{classKNN}, Support Vector Machines (SVMs) \cite{classSVM}, and even a core-generator for generating classifiers in FPGA \cite{classCORE}.}

{ To remain consistent with previous work \cite{HMAX2007} and provide a fair comparison, a boosting classifier was used when performing binary classification, and a linear (SVM) classifier was used when performing multi-class classification. The use of linear SVM is further supported by Misaki~\emph{et al.} \cite{FMRIclassifiers}, who did a comparison of multivariate classifiers in a visual object discrimination task using FMRI data from early stages of human visual and inferior temporal cortex.  Linear classifiers were found to perform better than non-linear classifiers, which they note is consistent with previous similar investigations \cite{FMRIclassifiersSupport1, FMRIclassifiersSupport2}. Misaki~\emph{et al.} also note that non-linear classifiers may perform better if larger datasets are used for training, or if fewer features are used. Non-linear classifiers can better fit the training data, but this comes with the risk of overfitting the classifier to the data, which is particularly problematic when only a few training samples are used.
}

The rest of this paper describes how the original model \cite{HMAX2007} was adapted for implementation on an FPGA to increase throughput and how these adaptations affect recognition accuracy. To test the FPGA implementation we performed a binary classification task on popular categories from the commonly referenced, publicly available Caltech 101 \cite{Caltech101} dataset as well as a tougher minaret dataset comprised of images downloaded from Flickr. We also investigated multi-class classification accuracy using Caltech 101.  Results are compared to previously-published test results on the same dataset using a software implementation of the HMAX model \cite{HMAX2007}. An analysis of how the image throughput rate and required hardware would change with input image size is also presented. The aim of this paper is not to beat the state of the art in terms of recognition accuracy, but rather to show how a given model can be adapted for implementation on an FPGA to drastically increase throughput while maintaining the same level of recognition accuracy.

\section{Original Model Description}
The version of the HMAX model used \cite{HMAX2007} has two main stages, each consisting of a simple and complex substage. We will call these Simple-1 (S1), Complex-1 (C1), Simple-2 (S2) and Complex-2 (C2) as is done in the original paper.

\subsection{S1}
In S1 the image is filtered at each location with Gabor filters applied at 16 different scales with the side length of a filter ranging from 7 to 37 pixels in increments of 2 pixels as shown in Table~\ref{Table:parameters}.
For each filter size the filter is applied at four different orientations (0$^o$, 45$^o$, 90$^o$, and 135$^o$). For each filter position the underlying image region is normalized before filtering to increase illumination invariance. The output of S1 consists of 64 filtered versions of the original image (16 scales $\times$ 4 orientations). The sign of the result is dropped and only the magnitude is passed to C1.  

\begin{table}
\caption{Parameters used in FPGA implementation of HMAX. This table is adapted from the parameter table shown in \cite{HMAX2007}.}
\begin{center}
\begin{tabular}{|c||c|c|c|c|}
\hline

\multirow{2}{*}{\textbf{Size Band\#}} & Subsampling & Filter Sizes & \multirow{2}{*}{$\sigma$} & \multirow{2}{*}{$\lambda$}\\
 & period $\Delta$  & $\diameter \times \diameter$ & & \\                                          \hline

\multirow{2}{*}{\textbf{Band 1}}& \multirow{2}{*}{4} & 7$\times$7 &1.3 &3.9\\
& & 9$\times$9 & 1.7 & 5.0 \\  \hline

\multirow{2}{*}{\textbf{Band 2}}&\multirow{2}{*}{5} &11$\times$11 &2.1 &6.2\\
&&13$\times$13& 2.5 & 7.4 \\ \hline

\multirow{2}{*}{\textbf{Band 3}}&\multirow{2}{*}{6} &15$\times$15 &2.9 &8.7\\
&& 17$\times$17 &3.3 & 10.0\\ \hline

\multirow{2}{*}{\textbf{Band 4}}&\multirow{2}{*}{7} &19$\times$19 &3.8 &11.3\\
&& 21$\times$21 &4.2 & 12.7\\ \hline

\multirow{2}{*}{\textbf{Band 5}}&\multirow{2}{*}{8} &23$\times$23 &4.7 &14.1\\
&&25$\times$25 & 5.2 &15.5\\ \hline

\multirow{2}{*}{\textbf{Band 6}}&\multirow{2}{*}{9} &27$\times$27 &5.7 &17.0\\
&&29$\times$29 & 6.2 &18.5\\ \hline

\multirow{2}{*}{\textbf{Band 7}}&\multirow{2}{*}{10}&31$\times$31&6.7&20.1\\
&&33$\times$33 & 7.2 &21.7\\ \hline

\multirow{2}{*}{\textbf{Band 8}}&\multirow{2}{*}{11}&35$\times$35 &7.8 &23.3 \\
&&37$\times$37 & 8.3 &25.0\\ \hline

\end{tabular}\label{Table:parameters}
\end{center}
\end{table}

\subsection{C1}
Filter responses are grouped by filter sizes into 8 size-bands as shown in Table~\ref{Table:parameters}. Within each size-band the response of a C1 unit is the maximum of the S1 units in that size-band over a small local spatial region ($2\Delta\times2\Delta$ from Table~\ref{Table:parameters}). The result is then subsampled (every $\Delta$ pixels) and output to S2. The output is therefore 32 sets of C1 units (8 size-bands $\times$ 4 orientations).

\subsection{S2}
S2 units have as their inputs C1 units from all four orientations. They compute the Euclidean distance between a predefined patch and the C1 units at every location. The patch sizes are $4\times4\times4$, $8\times8\times4$, $12\times12\times4$ and $16\times16\times4$ ($x \times y \times orientation$). For every S2 unit the patch distance is computed at every (x,y) location within every size-band and passed to C2.

\subsection{C2}
The C2 layer computes the minimum of the S2 distance for each patch across all locations in all size-bands. The number of C2 outputs is therefore equal to the number of S2 patches used.

\subsection{Classification}
Classification is performed directly on the C2 outputs. The choice of classifier can vary based on the required task. Previous work \cite{HMAX2007} presented results using a boosting classifier for binary classification, and a linear SVM one-vs-all classifier for multi-class classification.

\section{FPGA implementation}
\label{section:FPGA implementation}
\subsection{Hardware Description}

The large number of Multiply ACcumulate (MAC) operations required to implement the 64 filters in S1 and the 1000 patches in S2 make the number of multipliers available on an FPGA one of the limiting constraints for throughput. The second limiting constraint is the amount of internal memory available. We need to ensure we have enough memory to store all intermediate results, S2 patches, and S1 filters since we can save time by not loading S1 filters and S2 patches from external memory, as will be shown in Section~\ref{section:S2Hardware}. Multiple block RAMs are used in parallel whenever data wider than 16 bits needs to be stored. We chose to use the Xilinx XC6VLX240T from the Virtex 6 family for its large number of multipliers (768) combined with its reasonable price of \$1800 for a development board (Xilinx `EK-V6-ML605-G' board). The S1, C1, S2, and C2 stages were each implemented as separate modules in VHDL using a pipelined architecture.

\subsection{Edge Effects}
The most obvious way to speed up the model is to not waste resources on unnecessary computation. For this reason we chose to only compute filter responses and patch distances when the filter (S1) or patch (S2) has full support. We effectively ignored any computation which involves regions beyond the image edges. 

\subsection{S1 Filters}


\begin{figure*}
\centering
\includegraphics[width=0.8\textwidth]{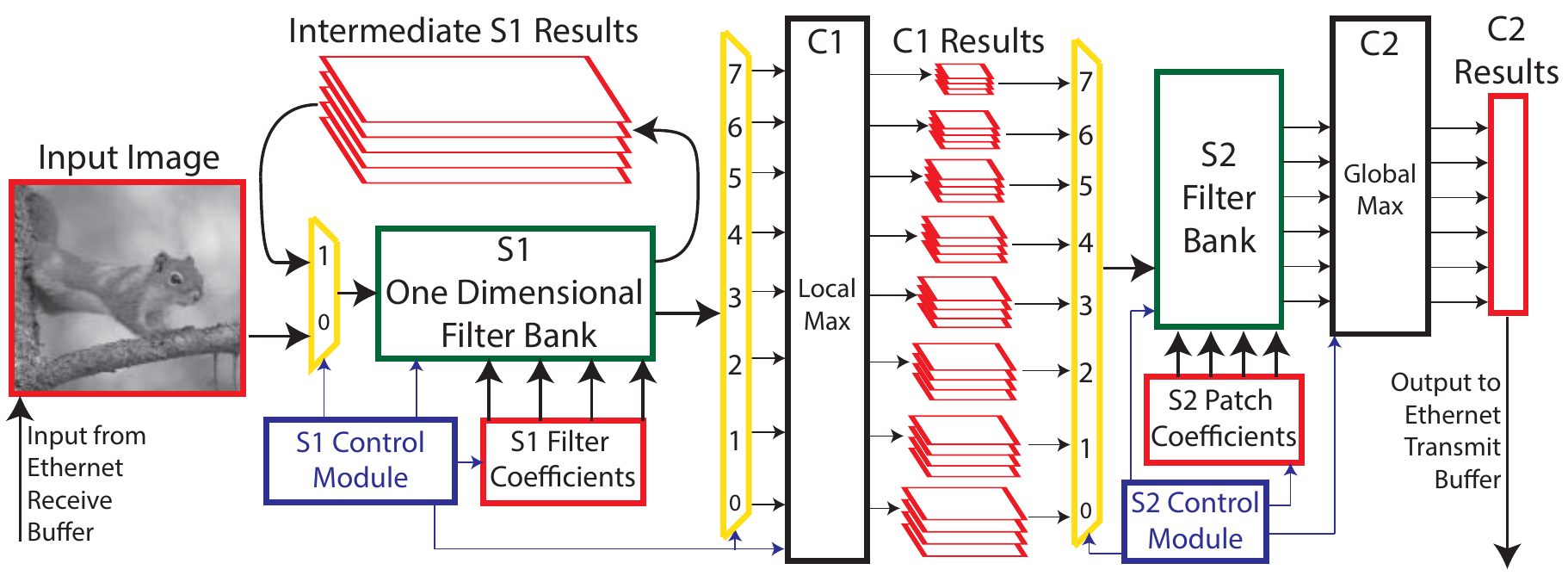}
\caption{Block diagram of our hardware implementation of HMAX. Red rectangles indicate the usage of block RAM. Input to the 1 dimensional S1 filters can come from either the input image or from RAM holding intermediate results. S1 results are sent to C1 where the maximum is computed over a local region (see Table~\ref{Table:parameters}) and stored. Each size band has its own dedicated RAM. A demultiplixer controls reading of C1 results for the S2 stage. C2 computes the global maximum of S2 outputs and stores the results in RAM before transferring them to the Ethernet transmit buffer.}\label{fig:HardwareImplementation}
\end{figure*}

The S1 layer consists of directionally selective Gabor receptive fields, similar to the selectivity of simple cells found by Hubel and Weisel \cite{HUBELWeisel} in V1. We implement cells at four different orientations (0$^o$, 45$^o$, 90$^o$ and 135$^o$) as was done in the original model \cite{HMAX2007}. Due to symmetry, we need not compute cells at orientations at or above 180$^o$. Each orientation is implemented at sixteen different scales and at every location in the image where full support is available. The equations defining the filters used in the original HMAX model \cite{HMAX2007} are repeated in \eqref{eq:Original Filters} for convenience. The equations for the filters are the product of a cosine function and a Gaussian weighted envelope:

\begin{equation}
\label{eq:Original Filters}
\begin{array}{l l}
F_{\theta}(x,y) &= e^{(-\frac{x_0^2+\gamma^2y_0^2}{2\sigma^2})} \times \cos{(\frac{2\pi}{\lambda}x_0)}\\[8pt]
x_0 &= x\cos{\theta} + y\sin{\theta}\\[8pt]
y_0 &= -x\sin{\theta} + y\cos{\theta}.\\
\end{array}
\end{equation}

Here $\lambda$ determines the spatial frequency at the filter's peak response, $\sigma$ specifies the radius of the Gaussian window and $\gamma$ squeezes or stretches the Gaussian window in the $y_0$ direction to create an elliptical window. For the 0$^o$, and 90$^o$ cases we can easily rewrite this equation as product of two separate functions as shown in \eqref{eq:Separable Filters}. The 45$^o$, and 135$^o$ terms are not separable unless we change the Gaussian weighting function to an isotropic function by specifying $\gamma~=~1$. By doing this we arrive at the equations for the 45$^o$ and 135$^o$ filters shown below:

\begin{equation}
\label{eq:Separable Filters}
\begin{array}{l l}
F_{0}(x,y) = E(x,y)*G(x,y)^T\\[8pt]
F_{90}(x,y) = E(x,y)^T*G(x,y)\\[8pt]
F_{45}(x,y) = E(x,y)*E(x,y)^T + O(x,y)*O(x,y)^T\\[8pt]
F_{135}(x,y) = E(x,y)*E(x,y)^T - O(x,y)*O(x,y)^T\\[8pt]
E(x,y) = e^{(\frac{-x^2}{2\sigma^2})}\cos{(\frac{2\pi x}{\lambda})}\\[8pt]
G(x,y) = e^{(\frac{-\gamma^2x^2}{2\sigma^2})}\\[8pt]
O(x,y) = e^{(\frac{-x^2}{2\sigma^2})}\sin{(\frac{2\pi x}{\lambda})}\\[8pt]
\end{array}
\end{equation}

Here $(x,y)$ is the location of the kernel value within the filter,  $O(x,y)$ is an odd Gabor filter, $E(x,y)$ is an even Gabor filter, and $G(x,y)$ is a pure Gaussian filter. $A*B$ designates the convolution of $A$ and $B$, while $A^T$ designates the transpose of $A$. By writing the filters in a separable manner, we can implement them using two passes of a one dimensional filter rather than one pass of a two dimensional filter \cite{antoniou2000digital}. The number of MAC operations required to implement a separable filter grows linearly with the side length of the filter rather than as the square of the side length and therefore results in a significant speed up, or in the case of FPGA implementation, a significant saving of resources. If we consider the specific case of implementing the 64 S1 filters at a single image location, we can compute the number of multiply accumulates required using
\begin{equation}
\label{eq:separable cost}
\begin{array}{l l l}
MAC_{original} &= 4 \times \sum_{i=1}^{16} [\diameter(j)^2] &= 36416\\
\\
MAC_{separable} &= 4 \times \sum_{i=1}^{16} [2\times \diameter(j)] &= 2816\\
\end{array}
\end{equation}
where $\diameter(j)$ is the side length of filter $j$ as indicated in Table~\ref{Table:parameters} and in \eqref{eq:Parameterise}.

Using separable filters reduces the number of required multiply accumulates from 36416 down to 2816, a reduction to less than 8\% of the originally required computation. Furthermore, each one-dimensional filter used has either even or odd symmetry about the origin, allowing us to sum values in the filter support either side of the origin before performing multiplication. By exploiting the symmetry of the filter the required multiplications are reduced by a further 50\%, freeing up more dedicated hardware multipliers for use in the more computationally intensive S2 stage of processing. Using separable instead of non-separable filters reduces the time taken to compute the S1 filter responses from 2.3 seconds to 0.3 seconds per 128$\times$128 image in Matlab.

To increase illumination invariance, the filter response at each location is normalized by the $l^2$ norm of its support, as is done in the original model. This normalization ensures that filters capture information about the local contrast and are unaffected by the absolute brightness of a pixel region. The $l^2$ norm is computed by first summing the squares in the x-direction, then summing the result in the y-direction and taking the square root. We timed this result to be available simultaneously with the filter results so that we can immediately perform division without the need to store intermediate results. Responses for filters at all four orientations are computed in parallel, eliminating the need to recompute or store the $l^2$ norm of the filter support for each orientation.

The filter kernels are all pre-computed and stored in a look up table (see Fig.~\ref{fig:HardwareImplementation}). Each filter is modified to have zero mean and an $l^2$ norm of $(2^{16}-1)$ to ensure that results are always less than 16 bits wide. The parameters used for these separable filters is shown in Table~\ref{Table:parameters}. These parameters can be written into equations as shown in \eqref{eq:Parameterise} below.

\begin{equation}
\label{eq:Parameterise}
\begin{array}{l l}
\diameter(j) &= 5 + 2\times j\\[8pt]
\Delta(b) &= 3+b\\[8pt]
\kappa(k) &= (4\times k)^2\\
\end{array}
\end{equation}
where $j$ is an index for filter sizes arranged from the smallest to largest (1 to 16). The diameter of filter $j$ is $\diameter(j)$. The filter is actually square with side length $\diameter(j)$ to avoid the complexity of implementing a round filter. The subsampling period of size band $b$ is written $\Delta(b)$. $k$ is an index for the size of patches (1 to 4 for the four different patch sizes). At each orientation a patch of size index $k$ will have size $\kappa(k)$.

\subsection{C1}
The C1 layer requires finding the maximum S1 response over a region of $2\Delta\times2\Delta$ and subsampling every $\Delta$ pixels in both $x$ and $y$ (for values of $\Delta$ see Table~\ref{Table:parameters}). We computed the maximum of a $2\Delta\times2\Delta$ region by first computing the maximum over adjacent non-overlapping regions of size $\Delta\times\Delta$. By taking the maximum across every 4 adjacent $\Delta\times\Delta$ regions we obtained the maximum over a $2\Delta\times2\Delta$ region, subsampled every $\Delta$ pixels in both $x$ and $y$.

Computing on data as it streams from S1 eliminates the need to store non-maximal S1 results (see Fig.~\ref{fig:HardwareImplementation}). As with the S1 layer, computation in C1 is performed on all four orientations in parallel. Each time C1 finishes computing the results for a size band, a flag is set which indicates to S2 that it can begin computation on that size band.


\subsection{S2}
\label{section:S2Hardware}
Even though the data coming into S2 has already been reduced by taking the maximum across a local pool and subsampling in C1, the S2 layer is where most of the computation takes place. The number of MAC operations required to compute all patch responses at a single location in the original model is:
\begin{equation}
\label{eq:patchMAC}
\begin{array}{l l}
250\times4\times\sum_{k=1}^4\kappa(k) &= 480 000\\
\end{array}
\end{equation}
where there are 250 patches per size and 4 orientations per patch, each of size $\kappa(k)$, which was defined in \eqref{eq:Parameterise}. The computation of these patch responses must be repeated at all locations within all size-bands.

We decided to use 1280 patches (320 per size) which was a compromise between speed of implementation and the number of patches. As in the original model, S2 patches are obtained from previously computed C1 results on images from both the positive and negative classes. Since S2 patches are simply portions of previously computed C1 outputs, the number of bits required to store each patch coefficient is 16. The closeness of a patch to a C1 region is computed as the Euclidean distance between the patch and that region.

We computed patch responses starting with the smallest sized patches ($x\times y\times orientation \rightarrow 4\times4\times4$) and computing their response at a single location. We then repeat this computation for all locations in the current size band, before moving onto the next patch size. Once all patch sizes have been computed for all locations in the current image size-band we move onto the next size-band as soon as it is available from C1. All patches of the size currently being considered are computed in parallel. Furthermore, the response at two different orientations is considered in parallel. This results in $320\times2 = 640$ parallel multiply-accumulate operations every clock cycle. This uses 640 multipliers and requires that 640 patch coefficients be read every clock cycle. Patch coefficients are stored in the FPGA's internal block RAM since the bandwidth to external RAM would not allow such high datarates. Using external RAM would require a data rate of $640\times16 bits\times100MHz~=~1Tb/s$ for a 100MHz clock.


\subsection{C2}
C2 simply consists of a running minimum for each S2 patch, computed by comparing new S2 results with the previously stored S2 minimum. This is performed for all 320 S2 patches of the current size simultaneously (see Fig.~\ref{fig:HardwareImplementation}). 

\subsection{Classifier}
\label{section:Classifier}
\begin{figure}
\centering
\includegraphics[width=0.4\textwidth]{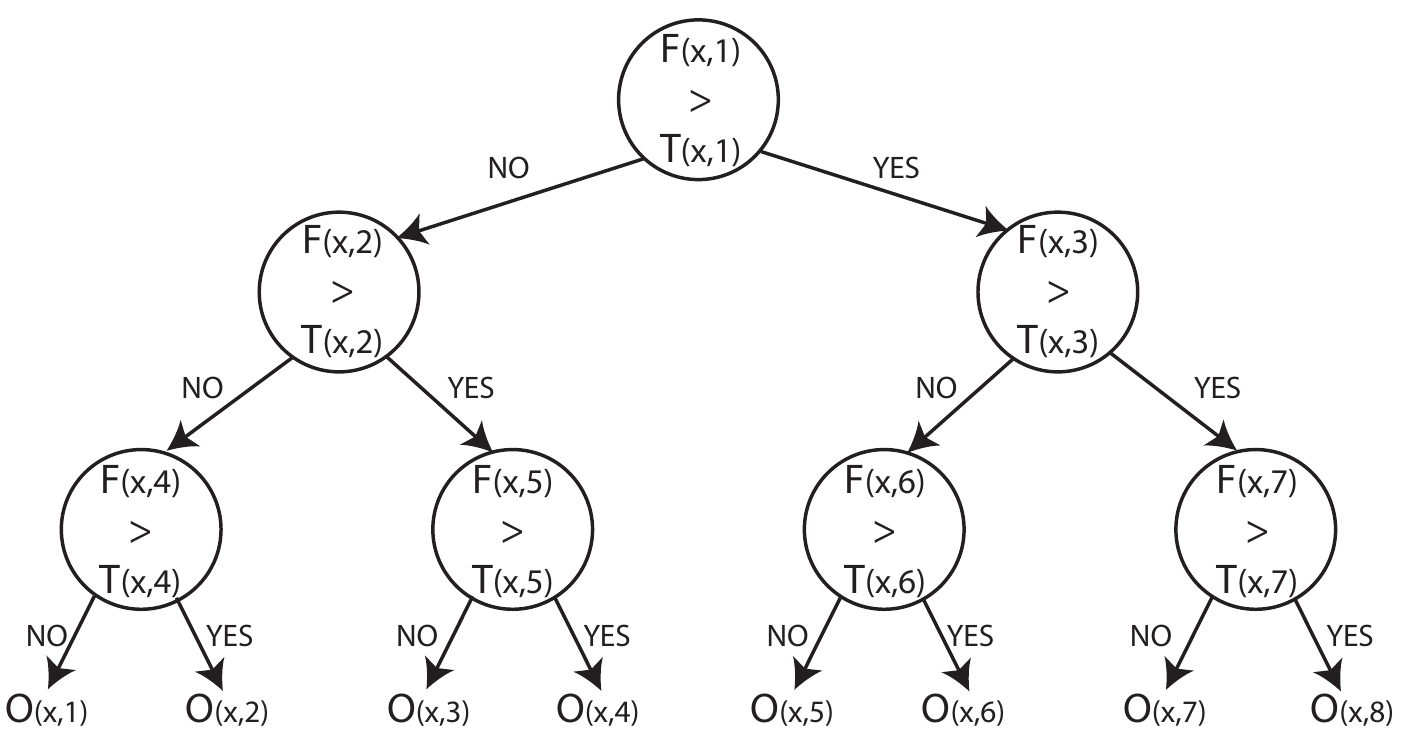}
\caption{A weak learner used in the gentle boosting algorithm. Each weak learner is a tree consisting of 7 nodes. $F(x,y)$ represents the feature used at node $y$ in weak learner $x$. $O_{(x,1)}$ through $O_{(x,8)}$ are the binary outputs of classifier $x$. Each output is a binary value 1 or -1.}\label{fig:WeakLearner}
\end{figure}

Results from \cite{HMAX2007} suggest that a boosting classifier is better than SVM for the binary classification problem. We used the gentleboosting algorithm \cite{GentleBoost} with weak learners consisting of tree classifiers each with a maximum of three decision branches before reaching a result as shown in Fig.~\ref{fig:WeakLearner}. We used 1280 weak learners in the classifier, each computed in series.

For multi-class classification a linear one-vs-all SVM classifier was chosen \cite{SVMmulticlassTraining, SVMmulticlass}. This is a simple linear classifier, but is memory intensive in its requirement for storing coefficients, as is discussed in Section~\ref{section:scalabilityHardware}.

In order to not restrict the FPGA implementation to only binary problems or only multi-class problems, the classifier was implemented separately on a host PC.


\subsection{Scheduling}
\label{section:Scheduling}
The FPGA implementation has an input FIFO buffer capable of holding up to four complete 128$\times$128 pixel images. As soon as at least one full image has been loaded into the buffer S1 will read the image. S1 then computes responses at all four orientations for the smallest filter simultaneously and outputs the results in a streaming fashion to C1. After computing the responses from the smallest filter, S1 filters will read in coefficients for the next filter size and compute the new filter responses. S1 will continue in this manner until responses for all filter sizes have been computed. S1 will read a new image from the input buffer as soon as it has completed the first pass with the largest separable filter, or as soon as an image becomes available if none are available at the time.

The C1 and C2 layers operate on the results of S1 and S2 as they are output in a streaming fashion during computation, thereby reducing the internal memory required to store intermediate results. This approach also ensures that C1 and C2 only add a negligible amount of processing time to the algorithm (less than 100 $\mu$seconds for an entire image).

{
Each stage (S1, C1, S2, C2) uses its own dedicated FPGA resources, thereby allowing all stages to run simultaneously.} Sharing of memory occurs between C1 and S2, where access is managed by setting and clearing flags. There is a separate memory unit and flag for each image band. When a flag is low, C1 has exclusive read/write access to the corresponding memory unit. Once C1 has finished storing results in the memory unit, it will set the corresponding flag high. When a flag is high, the S2 stage has exclusive read/write access to the corresponding memory block and will clear the flag once it has finished processing all data from that memory block, thereby transferring control back to C1.

If waiting for access to a particular memory block, a stage (C1 or S2) will begin processing as soon as access is granted (the very next clock cycle). Since results for each image band are stored separately, the S1 and C1 stages can process the next image band (and loop around) without having to wait. This allows S1 and C1 to be almost an entire image ahead in computation than the S2 stage, which is important because although the S1 and C1 stages take the same length of time to process each image band, the time taken by S2 varies. The S2 stage takes longer to compute on smaller image bands because their higher frequency of subsampling produces more C1 results on which computation must be performed (see Table~\ref{Table:parameters}). 
Buffering of C1 outputs in the manner described allows us to focus on matching the throughput of the S1 and C1 stage with the \emph{average} throughput (across image bands) of the S2 stage, without being troubled by how computation time in S2 varies with each image band.

S1 will not compute new results for an image band if the current results for that image band (from the previous image) have not yet been processed by S2 (i.e. if the relevant memory flag is still high). S1 will however still perform the first pass with a separable filter in the meanwhile to ensure it can start outputting results as soon as the flag is cleared.

{ Results from S2 stream to C2, which writes the final results to an output buffer for communication back to the host PC.}

\section{Scalability of FPGA implementation}
In this section we show how the input image size affects the hardware resources and time required for computation using the FPGA implementation described in Section~\ref{section:FPGA implementation}. The described FPGA implementation was specifically designed to operate on images of size 128$\times$128 pixels and is therefore not necessarily recommended as the best implementation for larger or smaller images. Nevertheless, if implementing a new design to operate on larger (or smaller) images, extrapolating the current design to different sizes provides a good starting point.

\subsection{Hardware Resources}
\label{section:scalabilityHardware}
The number of bits in the counters used to track the progress of computation on the input image and intermediate results in stages S1, C1, and S2 will need to increase to handle larger images. This increase scales as:

\begin{equation}
\label{eq:counter size}
\begin{array}{l l}
Counter Bits \propto \log_2{\sqrt{N}}\\
\end{array}
\end{equation}

\noindent where $N$ is the number of pixels in the input image and the image is assumed to be square, having side length $\sqrt{N}$. This increase in required hardware is negligible, especially in comparison to the increase in internal RAM required to store the input image and intermediate results in the S1 and C1 stages.
The internal RAM requirement scales proportionally to $N$ for large images. Due to the nature of computation in S2 and C2, no additional RAM is required in those stages when the image size increases. The number of elements required to compute multiplication, addition, division, and square roots remains unchanged in all stages. The total required internal RAM is the sum of the RAM required by all stages.

Internal RAM is required for three purposes in S1: storing the input image, storing intermediate results between the first and second passing of the separable filter and finally, to store the S1 filter coefficients. The required RAM can be explicitly calculated using \eqref{eq:S1bits} below.

\begin{equation}
\label{eq:S1bits}
\begin{array}{l l}
S1_{bits} &= S1_{input} + S1_{intermediate}+ S1_{filters}\\[8pt]
S1_{input} &= 4\times N \times 8\\[8pt]
S1_{intermediate} &= 5\times N \times 23\\[8pt]
S1_{filters} &= \sum_{j = 1}^{16}{(2 \times(3+j) \times 16)}\\[8pt]
\end{array}
\end{equation}

\noindent $N$ represents the number of pixels in the input image. The $input$ buffer has to hold four images (a FIFO buffer) with 8 bits per pixel. The $intermediate$ results require 5 buffers (one for each orientation and one for calculating the $l^2$ norm of the filter support). Each result consists of 23 bits. For storage of the $filters$, the $j^{th}$ filter (ordered smallest to largest) consists of 2 separable filters, each with $(3+j)$ coefficients and 16 bits per coefficient.

The output of the S1 stage does not require RAM for storage since each result is processed by C1 as soon as it becomes available, but C1 does require RAM for intermediate and final results. The RAM required by C1 can be explicitly calculated using \eqref{eq:C1bits} below.

\begin{equation}
\label{eq:C1bits}
\begin{array}{l l}
C1_{bits} &= \sum_{b = 1}^{8}C1_{size}(b)\times 16\\
\\
C1_{size}(b) &= \frac{S1_{size}(b)}{\Delta(b)^2}\\
\\
S1_{size}(b) &= 4\times(\sqrt{N}-\diameter(2b)+1)^2\\
\end{array}
\end{equation}

\noindent The number of valid S1 results in image band $b$ is then given by $S1_{size}(b)$, where $\diameter(2b)$ was previously defined in \eqref{eq:Parameterise} and there are 4 orientations. The number of C1 results can then be calculated knowing the number of S1 results and the subsampling period $\Delta(b)$, which was also previously defined in \eqref{eq:Parameterise}. Each C1 result occupies 16 bits.

The RAM required for S2 is constant across image sizes and can be written explicitly as:

\begin{equation}
\label{eq:S2bits}
\begin{array}{l l}
S2_{bits} &= \sum_{k = 1}^{4}320\times 4\times \kappa(k) \times 16 \\
\end{array}
\end{equation}
where $k$ is an index of patch size. There are 320 patches per size and 4 orientations per patch, each with $\kappa(k)$ coefficients as previously defined in \eqref{eq:Parameterise}. Each coefficient occupies 16 bits.

C2 requires only enough RAM to hold the final C2 results.

\begin{equation}
\label{eq:C2bits}
\begin{array}{l l}
C2_{bits} &= 1280\times42\\
\end{array}
\end{equation}

\noindent where there are 1280 C2 features each consisting of 42 bits.

Although we implement the classifier on the host PC, it is possible to determine the resources required by the classifier. {The most memory intensive classifier used in this paper is the 102 class one-vs-all linear SVM classifier, for which the memory requirements are:}
\begin{equation}
\label{eq:ClassifierBits}
\begin{array}{l l}
Classifier_{bits} &= 102\times1280\times32 + 84\\
 &= 4178004 bits\\
\end{array}
\end{equation}
where there are 102 possible classes, 1280 C2 features, 32 bits per coefficient, and up to 84 bits required to hold the result. The current FPGA implementation does not have enough remaining internal memory to hold all these coefficients, but the coefficients could easily fit into external RAM, or the classifier could be run on a second FPGA. If running at 190 images per second, an external memory bandwidth of $102\times1280\times32\times190 = 794Mbps$ per second would be required, which is only about 6\% of the available 12.8Gbps bandwidth on the targeted FPGA platform. In our implementation, running the classifier on a host PC did not affect the system throughput.


\subsection{Time}
\label{section:scalabilityTime}
The time taken to process an image is dominated by the S1 and S2 stages. The C1 and C2 stages perform simple maximum operations on each valid data point as it becomes available and therefore do not contribute significantly to the time taken to process an image. {The time computed in the equations below is in units of clock cycles and the actual time taken for computation therefore depends on the FPGA clock frequency.}

The time taken to compute S1 can be accurately approximated as the time required to do 2 passes of the image for each of the 16 separable filter sizes \eqref{eq:S1Time}. All four orientations are simultaneously computed in parallel and therefore the multiple orientations do not add to computation time.

\begin{equation}
\label{eq:S1Time}
\begin{array}{l l}
S1_{time} &= 2\times N \times 16\\
\end{array}
\end{equation}

\noindent where $S1_{time}$ is in units of clock cycles, $N$ is the number of pixels per image and 16 filter sizes are implemented.

In S2, all 320 patches of the same size are considered simultaneously and within each patch, computation is performed at two orientations simultaneously.

\begin{equation}
\label{eq:S2Time}
\begin{array}{l l}
S2_{time} = \sum_{b = 1}^{8}\sum_{k = 1}^{4}S2_{size}(b,k) \times \kappa(k) \times2\\
\\
S2_{size}(b,k) = (\sqrt{C1_{size}(b)}-\sqrt{\kappa(k)} +1)^2\\ 
\\
\end{array}
\end{equation}

\noindent where $S2_{size}(b,k)$ is the number of valid S2 results for size band $b$ and patch size index $k$. $S2_{size}(b,k)$ is zero whenever the size of the C1 results is smaller than the patch size, that is when $C1_{size}(b)< \kappa(k)$. $\kappa(k)$ is the patch size and was previously defined in \eqref{eq:Parameterise}. $S2_{time}$ is the total time (in clock cycles) taken to compute all patch responses of all sizes in every size band.

{ If the multi-class one-vs-all linear SVM classifier were to be implemented on the FPGA with 102 classes and only a single hardware multiplier, the time taken could be computed as
\begin{equation}
\label{eq:classTime}
\begin{array}{l l}
Classifier_{time} &= 1280\times102\\
\end{array}
\end{equation}
for 1280 C2 features and 102 classes. The time taken for classification would not be dependent on the input image size. Using a single multiplier would enable a throughput of up to 765 images per second when using a 100MHz clock.}

\section{Simulation}
Four different sets of code were used in simulation. The first is a Matlab implementation of the HMAX model which was retrieved from HMAX website \cite{HMAXwebsite}. This was used as a benchmark against which to compare our modified implementation of HMAX for FPGA to verify that the modifications made did not severely compromise recognition accuracy. We refer to this original HMAX implementation as `HMAX CPU'.

The second, third, and fourth sets of code are Matlab, C++, and VHDL implementations respectively of our modified version of HMAX for FPGA. These implementations are functionally equivalent and we refer to them as `HMAX FPGA'. The Matlab code was used to make initial changes to the model and test accuracy on small datasets. Once satisfied with the changes made, a faster C++ implementation was written and used to verify the modified model on larger datasets. Finally, the actual VHDL code required to implement the proposed model in FPGA was written. This VHDL code was used to determine possible clock speeds and image throughput as well as to verify that the proposed FPGA model could be implemented using the resources available on the targeted FPGA platform (Xilinx Virtex 6 XC6VLX240T). Both final and intermediate results from the modified Matlab, C++, and VHDL codes were compared to verify that all three were performing the same computation.

\section{Hardware Validation}
The results of simulation were verified through implementation on the Xilinx Virtex 6 ML605 development board. A C++ interface was written for the host PC which handles Ethernet communications with the ML605 board and performs classification. The C++ code transmits four images to the ML605 board to fill the input buffer (described in Section~\ref{section:Scheduling}), then waits for all 1280 C2 values from an image to be returned before transmitting the next image. Reading of images from the hard drive and classification are both performed while waiting for the next set of C2 values from the FPGA, thereby adding negligibly to the overall computation time. Classification results are written to an output file as they are computed. For further verification C2 results from FPGA could be optionally written to disk for direct comparison against simulated C2 results.

\begin{figure}
\centering
\includegraphics[width=0.4\textwidth]{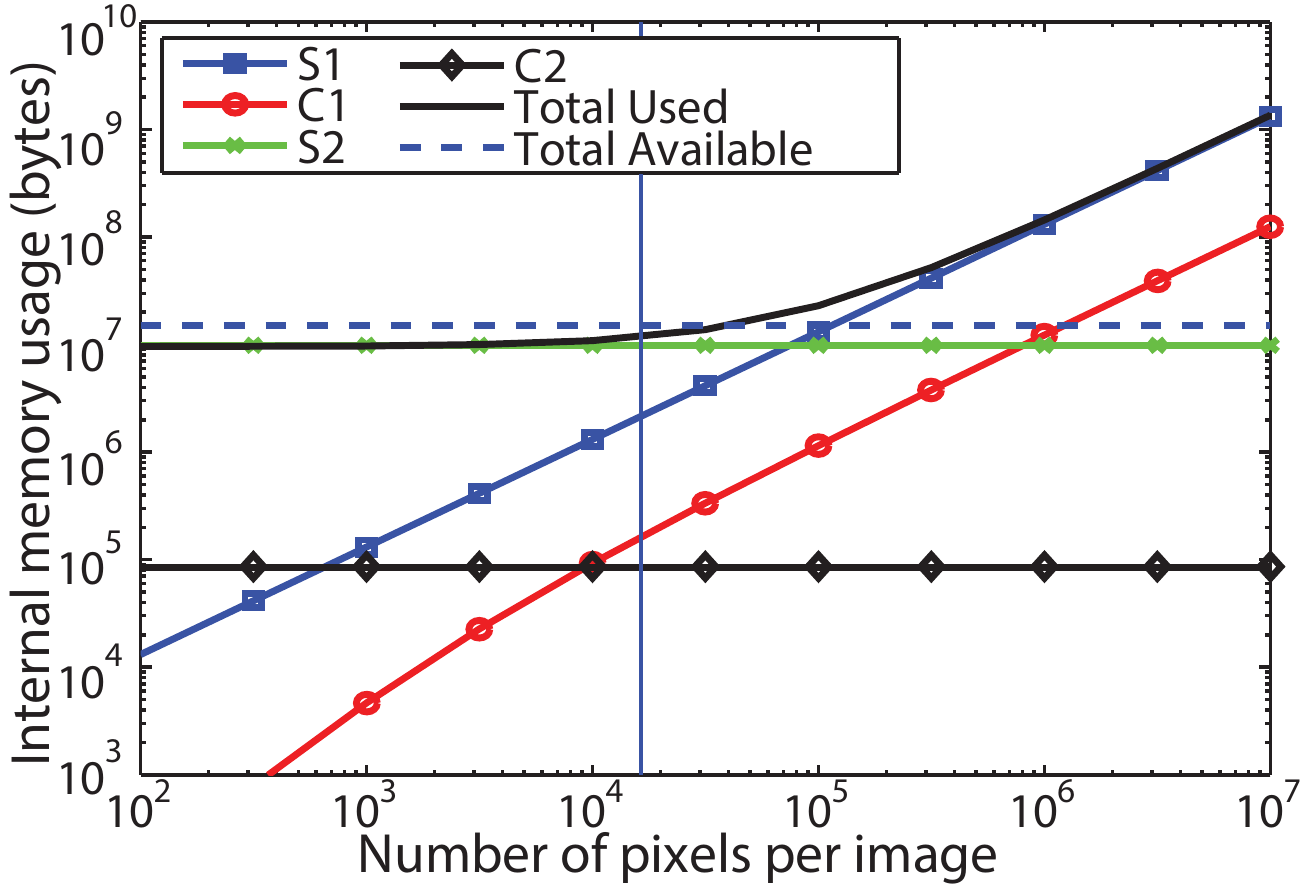}
\caption{Theoretical analysis of the internal RAM required by each stage as well as the total RAM required and total RAM available on the selected FPGA. The vertical line shows the number of pixels in a 128$\times$128 image, for which this implementation was designed.}\label{fig:ScaleMem}
\end{figure}

\section{Results}

\subsection{FPGA code analysis}
Using the Xilinx ISE, the VHDL code for implementing HMAX on FPGA was analyzed. For simplicity we use a single clock for all stages within the model. All lookup tables, S1 filters, and S2 patches as well as all intermediate results are stored in internal block RAM, as shown in Fig.~\ref{fig:HardwareImplementation}. The system has a latency of 600k clock cycles when processing a single image, but can maintain a throughput of an image every 526k clock cycles. Implementation of the full model indicates that the design can run at a clock frequency of 100MHz (10ns period). A 100MHz clock results in a latency of 6ms for processing a single image and a maximum throughput of 190 images per second when processing multiple images. These figures are achieved assuming that the input figure is a 128$\times$128 pixel 8-bit per pixel grayscale image.
The throughput of the design is determined by the throughput of the slowest stage in the pipeline. Computational resources should therefore be allocated in such a way that all stages have roughly the same throughput. This has been done as is evident in the distribution of multipliers between the S1 and S2 stages. S1 is the slowest stage, limiting the throughput to 190 images per second using 77 multipliers at 100MHz clock frequency, while S2 is capable of a throughput of 193 images per second, but uses 640 multipliers.

If we were to create an optimal implementation of S1 using non-separable filters with a 100MHz clock, then S1 alone would require over 1600 multipliers to achieve the same throughput of 190 images per second (unless a scale space approach was adopted). This is over double the number of hardware multipliers available on the chosen FPGA.

Table~\ref{table:FPGAResources} shows the total resources used by the HMAX implementation.

\begin{table}
\caption{FPGA resources used by HMAX}
\label{table:FPGAResources}
\begin{center}
\begin{tabular}{|l||c|c|c|}
\hline
\textbf{Resource} & \textbf{Used} & \textbf{Available} & \textbf{\% used} \\  \hline
\textbf{Multipliers} & \multirow{2}{*}{717} & \multirow{2}{*}{768} & \multirow{2}{*}{93}\\
\textbf{(DSP48E1)} & & &\\  \hline
\textbf{Internal RAM} & \multirow{2}{*}{373} & \multirow{2}{*}{416}& \multirow{2}{*}{89}\\
\textbf{(RAM36E1)} & & &\\  \hline
\multirow{2}{*}{\textbf{Slice Registers}} & \multirow{2}{*}{66 196} & \multirow{2}{*}{301 440}& \multirow{2}{*}{21}\\
 & & &\\  \hline
\multirow{2}{*}{\textbf{Slice LUTs}}  & \multirow{2}{*}{60 872} & \multirow{2}{*}{150 720}& \multirow{2}{*}{40}\\
 & & &\\  \hline
\end{tabular}
\end{center}
\end{table}

\subsection{Scalability}

Fig.~\ref{fig:ScaleMem} shows the internal RAM requirements computed using the equations presented in Section~\ref{section:scalabilityHardware}, as well as the total block RAM available on the selected Virtex 6 FPGA (14976kb, dashed line) and the image size for which the algorithm was designed (128$\times$128 pixels, vertical line).
Since all S2 patches of the same size are computed in parallel, the number of patches does not affect computation time, but will be limited by the number of available multipliers and amount of RAM available (see Table~\ref{table:FPGAResources}).

\begin{figure}
\centering
\includegraphics[width=0.4\textwidth]{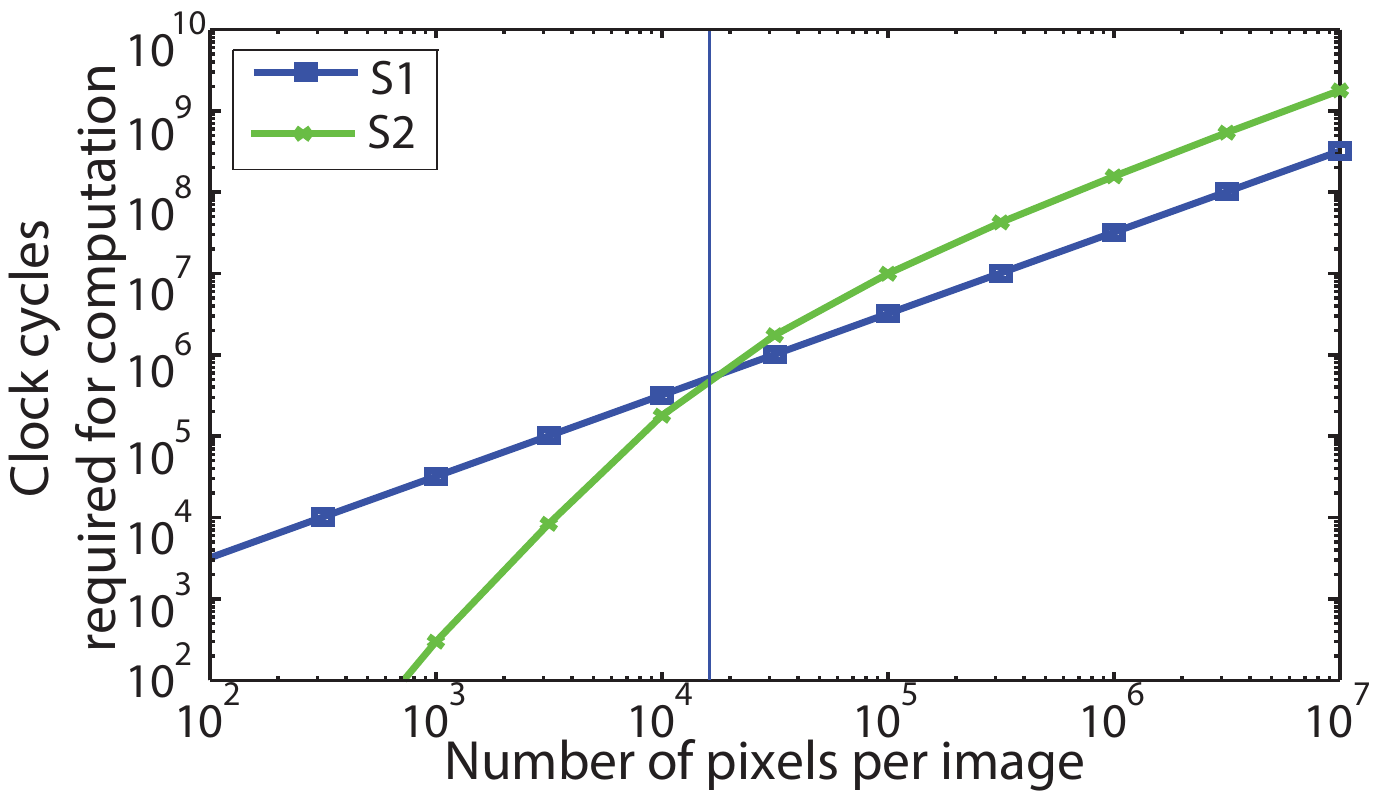}
\caption{A theoretical analysis of the time taken to compute each stage of HMAX in the current architecture. Due to the pipelined nature of the computation, the rate at which images can be processed is limited by the stage which takes the longest time. The vertical line shows the number of pixels in a 128$\times$128 image, for which this implementation was designed. The time required to compute the two longest stages is equal at this point as a result of the effort to allocate resources in such a way as to maximize throughput.}\label{fig:ScaleTime}
\end{figure}

The time taken to compute the S1 and S2 stages is shown in Fig.~\ref{fig:ScaleTime} along with the number of pixels for which the current implementation was designed (vertical line). The throughput of the complete system is limited to the throughput of the slowest stage.

The time taken to compute S2 can be seen as the time which would be taken to compute all results (even partial results on edges) minus the time which is saved by not computing edge results. The time saved by not computing at edges is significant at an image size of 128$\times$128. The time saved grows proportionally to the side length of the image $\sqrt{N}$, which is much slower than the time to compute all results (which grows linearly with $N$). This is why the time for S2 grows linearly with $N$ only for large $N$. S1 always grows linearly with $N$.

The design of the current framework ensures that the time taken for S1 and S2 is roughly equal (within 2\%) for images of size 128$\times$128, thereby ensuring that computational resources in each stage are not sitting idle waiting for the other stage to finish computing. If working with images of a different size, resources would ideally be reallocated to ensure that S1 and S2 still take equal time.

\subsection{Caltech 101 binary classification}
\begin{figure}
\centering
\includegraphics[width=0.45\textwidth]{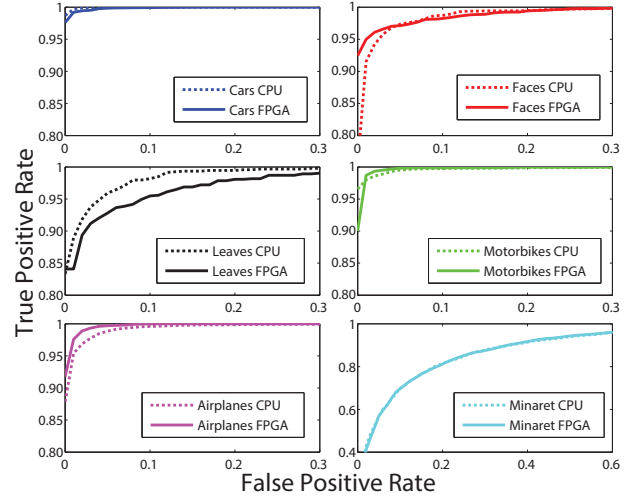}\\
\caption{Receiver Operating Characteristics for the binary classification task on Caltech 101 popular image categories and Minaret datasets. Each curve is the result of a mean over 10 trials. Note that the True Positive Rate axis is different for the Minaret classification task.}\label{fig:ROC}
\end{figure}

Two datasets were used to test the recognition accuracy of our modified HMAX model. The first is the often referenced Caltech 101 dataset \cite{Caltech101}. Recognition accuracy of popular categories in this dataset were presented for the HMAX model in \cite{HMAX2007}. We ran our own binary classification simulations on these categories using both the downloaded and modified versions of HMAX. {The binary task constituted discriminating the class in question (airplanes, cars, faces, leaves, or motorbikes) from the background class. In each case, half the images from the class in question and half images from the background class were used for training. The remaining images from both the class in question and the background class were used for testing.} In each case 10 trials were run. The accuracy reported in Table~\ref{table:Caltech101PopularCategories} is the percentage of correct classifications at the point on the ROC curve (Fig.~\ref{fig:ROC}) where the false positive and false negative rates are equal. Looking at the mean accuracy for this metric, the FPGA implementation achieves 0.24\% higher accuracy than the original CPU implementation. This shows that the modifications made for the FPGA implementation have not adversely affected recognition accuracy. 

\begin{table}
\caption{Comparison of recognition accuracies obtained from original HMAX code and FPGA implementation on popular categories in Caltech 101}
\label{table:Caltech101PopularCategories}
\begin{center}
\begin{tabular}{|@{ }c@{ }||@{ }c@{ }|@{ }c@{ }|@{ }c@{ }|}
\hline
\textbf{Category}   & HMAX \cite{HMAX2007}  & HMAX CPU      & HMAX FPGA \\  \hline
\textbf{Airplanes}  & 96.7                  & 97.1          & 98.2      \\  \hline
\textbf{Cars}       & 99.7                  & 99.3          & 99.2     \\  \hline
\textbf{Faces}      & 98.2                  & 95.8          & 96.4     \\  \hline
\textbf{Leaves}     & 97.0                  & 94.6          & 93.7      \\  \hline
\textbf{Motorbikes} & 98.0                  & 98.3          & 98.8      \\  \hline
\end{tabular}
\end{center}
\end{table}

\begin{figure}
\centering
\includegraphics[width=0.45\textwidth]{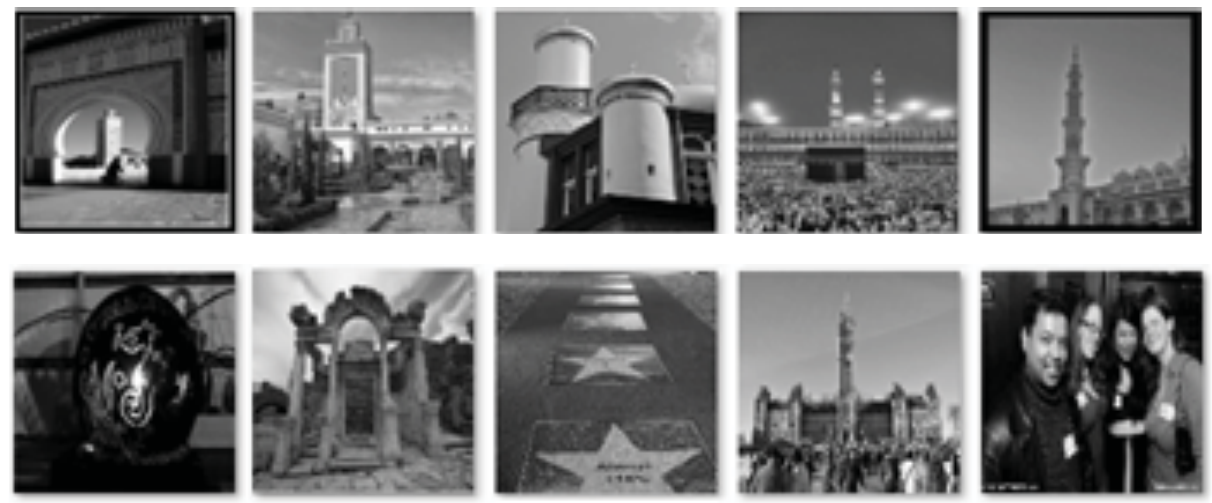}
\caption{A sample of images from the minaret (top~row) and background (bottom~row) classes used in the minaret binary classification task.}\label{fig:MinaretExamples}
\end{figure}

\subsection{Binary classification on Flickr dataset }
The binary minaret classification task was performed on a dataset containing 662 images of minarets and 1332 background images. The minaret (positive) images were obtained from Flicker by searching for ``Minaret" while negative images were obtained by periodically downloading the most recently uploaded Flicker image. Examples of these images are shown in Fig.~\ref{fig:MinaretExamples}. Ten random splits were used for classification and testing, with the test set consisting of 1000 negative and 500 positive images. The remaining images constitute the training set. This test was performed with both the downloaded HMAX code and the modified HMAX code for FPGA. The results are shown in Table~\ref{table:MinaretResults}. The metric used is the percentage of correct classifications at the point where false positive and false negative rates are equal. As expected, using 2000 features instead of 1280 improves the accuracy for both the CPU and FPGA implementations. The accuracy of the FPGA implementation is within 1\% of that of the original model.

\begin{table}
\caption{Comparison of results obtained from original HMAX code and FPGA implementation on minaret classification task}
\label{table:MinaretResults}
\begin{center}
\begin{tabular}{|@{ }c@{ }||@{ }c@{ }|@{ }c@{ }|@{ }c@{ }|@{ }c@{ }|}
\hline
\textbf{Model} & HMAX CPU & HMAX CPU & HMAX FPGA & HMAX FPGA\\  \hline
\textbf{Features} & 2000 & 1280 & 2000 & 1280\\  \hline
\textbf{Accuracy}&82.9& 82.2&82.2 & 81.3\\  \hline

\end{tabular}
\end{center}
\end{table}

\subsection{Caltech 101 multi-class one-vs-all}
\label{section:Caltech101Results}
\begin{figure}
\centering
\includegraphics[width=0.4\textwidth]{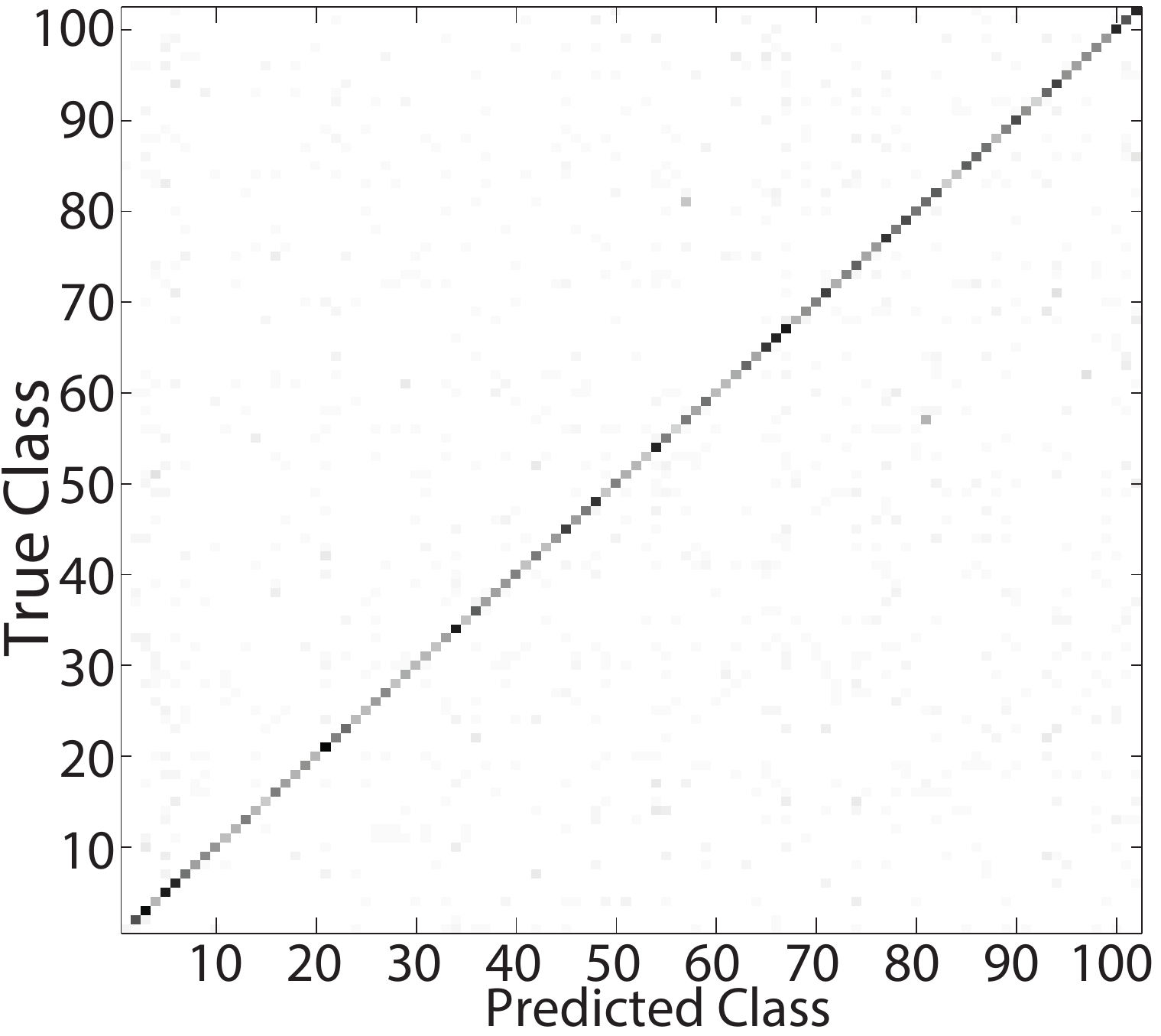}
\caption{Confusion matrix averaged over 10 trials for the 102 category multi-class one-vs-all test performed on the Caltech 101 database. The low accuracy on the extreme bottom left of the diagonal is the background category. The largest confusion is between the `schooner' (81) and `ketch' (57) categories, which are similar cases of sailboats.}\label{fig:ConfusionMatrix}
\end{figure}

A second test using the Caltech 101 database is the multi-class one-vs-all test. For this we used 15 training examples per category, as was done in \cite{HMAX2007}. Testing was performed using 50 examples per category or as many images as remained if fewer than 50 were available. Each of the categories was weighted such that it contributes equally to the result as was done in \cite{HMAX2007}. This is a 102 category problem including the background category. Using the one-vs-all linear SVM multi-class classifier from \cite{SVMmulticlass} we achieved a mean accuracy of 47.2 $\pm$ 1.0\% over 10 trials, which is in agreement with the result of 44 $\pm$ 1.14\% reported in \cite{HMAX2007} for the same task. The slight increase in accuracy can be attributed to the fact that our FPGA implementation uses 1280 features compared to 1000 features used in \cite{HMAX2007}. The confusion matrix for the 101 multi-class one-vs-all problem is shown in Fig.~\ref{fig:ConfusionMatrix}.

\subsection{Hardware Validation}
\label{section:testing}

\begin{table}
\caption{Maximum Throughput for Each Stage in Images/sec}
\label{table:MaxThroughput}
\begin{center}
\begin{tabular}{|l|c|c|c|c|c|}
\hline
\textbf{Stage}     &Input Buffer& S1   & C1  & S2  & C2     \\  \hline
\textbf{Throughput}&6100 &190   & 552 & 193 & 10 000  \\\hline
\end{tabular}
\end{center}
\end{table}

Results from analyzing VHDL code were verified by implementing the code on the ML605 board and processing the Caltech 101 database. The entire dataset consisting of 9144 images was processed ten times in different trials. The time taken to complete processing was measured from when the first image is read from disk until the last classification result is written to disk. The time taken to process the entire Caltech 101 database was measured as 48.12s $\pm 57\mu$s, which is a throughput of 190 images/sec and agrees with VHDL simulation predictions (shown in Table~\ref{table:MaxThroughput}) to within 0.01\%. Accuracy of the VHDL implementation was also verified against simulations. Both classification results and C2 outputs from testing were verified against simulation and found to exactly match.

\subsection{Comparison to other approaches}
To the best of our knowledge 190 images/sec is the fastest reported implementation of this version HMAX. Direct comparisons with other versions are not always straightforward because both the number of patches and their sizes can vary, as well as the size of the input image or even the model itself.

In 2010 Sedding \emph{et al.} \cite{CudaTimes2010} presented a time of 86.4ms for 4075 patches using custom code on an NVIDIA GeForce285 GTX. They used sparse features as proposed by Mutch and Lowe \cite{HMAXsparse} and claimed a shorter runtime than both the Feature Hierarchy Library (FHLib) \cite{HMAXsparse} and the GPU based Cortical Network Simulator (CNS) \cite{GPUhmaxCNS}.
In our aim to recreate the original model we chose not to use sparse features, but using sparse features would allow us either a 4$\times$ speedup or it would allow us to implement 4$\times$ as many patches at the same speed (resulting in 5120 patches) on the ML605 board. Their implementation also operates on larger images, with shortest side measuring 140 pixels. If our 1280 dense patch implementation was to run on an image measuring 140$\times$186 pixels (assuming a 3$\times$4 aspect ratio), it would still take under 12ms to complete.

On Caltech 101 with 15 training and 50 test samples per category, our 1280 patch 128$\times$128 pixel model achieves an accuracy of 47.2\% (see Section~\ref{section:Caltech101Results}) whereas Sedding \cite{CudaTimes2010} achieves 37\%, most likely a result of using sparse features. In terms of speed our implementation takes 5.3ms whereas theirs takes 86.4ms. They can reduce their processing time to 8.9ms if they only compute 240 patches, but this will come at the expense of even lower accuracy (less than 30\% on the same task).

\section{Discussion}
\label{section:Discussion}
The previous section shows that a massive increase in throughput can be achieved with almost no change in recognition accuracy. In this paper the aim has been to achieve a very high throughput as an argument for the use of FPGA in hierarchical models, but one could just as easily trade speed for accuracy. Interestingly our FPGA implementation of HMAX uses more S2 patches (1280) than the 1000 used in \cite{HMAX2007}. This increase in the number of patches was implemented simply because the additional resources required for the patches were available and the parallel processing of patches means that as long as resources are available, adding more patches does not affect throughput.

\begin{figure}
\centering
\includegraphics[width=0.4\textwidth]{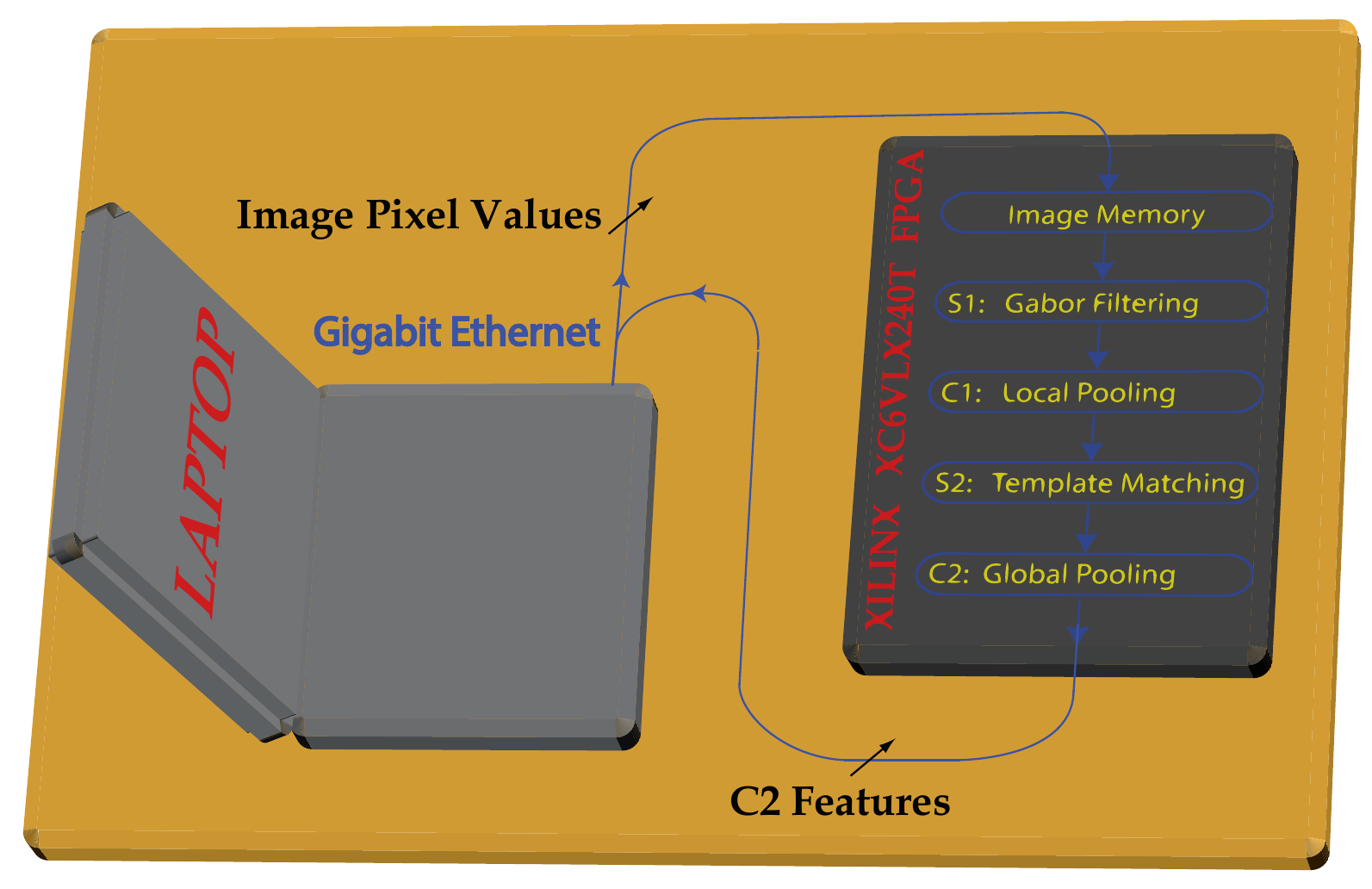}
\caption{An illustration of the portable hardware setup for the binary classification system showing a laptop communicating pixel values over gigabit Ethernet to a Xilinx ML605 evaluation board containing the Xilinx Virtex 6 XC6VLX240T FPGA on which the HMAX model runs. C2 features are returned to the laptop via the same gigabit Ethernet interface.}\label{fig:Block Diagram}
\end{figure}

The issues of image acquisition, rescaling and conversion to grayscale are not tackled by the current model since these will be application specific. The model requires that images are prescaled to 128$\times$128 pixels and converted to 8 bit grayscale before they are processed. The FPGA model requires an input image in the form of raw pixel values. For 190 images per second this translates to just over 3MB of data per second, which is well within the capabilities of the evaluation board's PCI express or gigabit Ethernet interfaces, as has been verified through testing in Section~\ref{section:testing}. If using a laptop, the system can run over gigabit Ethernet allowing it to be portable as shown in Fig.~\ref{fig:Block Diagram}.

The HMAX model used in this paper is one which was freely available in easy to follow Matlab code. It does not represent the least computationally intensive, or most accurate version of the HMAX model. The creators of the model are continuously working on improvements and a number of newer iterations have been presented \cite{HMAXsparse}. One of the most significant changes is the use of a scale-space approach such that the image is rescaled and reprocessed multiple times by filters of a single fixed size rather than keeping the image the same size and using multiple filters of varying size. Many recent implementations \cite{PENN1, PENN2, PENN3, PENN4, PENN5, PENN6} make use of 12 orientations instead of 4, which increases accuracy although it comes at the expense of extra computation time.

We achieved a key speedup in the S1 layer by exploiting the known structure of filters, which allowed us to implement the Gabor filters as separable. The unsupervised learning in S2 means that its structure is not known a priori. If the model were changed to S2 patches of a known structure which could be similarly exploited then further significant speedups could be achieved, but the effect on recognition accuracy would have to be further investigated.

Another change which greatly reduces computational complexity is the use of sparse S2 patches as proposed by Mutch and Lowe \cite{HMAXsparse}. In their model only the S1 orientation with maximal response is considered at each image location, thereby reducing the number of orientations in S2 from 4 to 1, which reduces the number of required multiply accumulates to only a quarter of the original. These sparse S2 features are used in most recent works \cite{PENN1, PENN2, PENN3, PENN4, PENN5, PENN6}. The effect on throughput of using sparse versus dense features, and of changing the number of orientations from 4 to 12, can be found in a recent paper by Park~\emph{et al.} \cite{PENN5}. Despite running on four FPGAs, each of which is more than twice as large as our FPGA (Virtex 6 SX475T versus LX240T), their dense implementation of HMAX using four orientations runs at roughly 45 images per second. However there are certain differences, they operate on larger images (256$\times$256 versus 128$\times$128), and use more patches (4075 versus 1280). Using four FPGAs, we could run four copies of our model in parallel, each with different patches, thereby giving us $1280\times4 = 5120$ patches while maintaining throughput of 190 images per second. We also use an equal number of patches of each size, whereas more recent approaches typically use more small (4$\times$4) and less large (16$\times$16) patches to reduce computation. To summarize in comparison with Park~\emph{et al.}, we could implement more patches (5120 versus 4075), with a higher percentage of large patches, and a 4$\times$ higher throughput if 4 FPGAs were used. Their implementation uses significantly larger FPGAs than ours (containing 2016 versus 768 multipliers), but also operates on 4$\times$ larger images, making a direct comparison difficult.

A common bottleneck for parallel architectures lies in the available bandwidth to memory and structuring how memory is accessed. For example, if two cores simultaneously request data from memory, one will have to wait for the other before it can access memory. In the presented FPGA implementation this was overcome by using the internal block RAM of the FPGA which resulted in a bandwidth of over 1 Terabit per second, which could be difficult to maintain on other platforms. Other implementations of HMAX which have recently been published also make use of internal block RAM to overcome this memory access bottleneck \cite{PENN1, PENN2, PENN3, PENN4, PENN5, PENN6}.


The size of the current filters and patches are designed to operate on small images. Even if higher resolution images are available, they should be rescaled to 128$\times$128 if they are to be processed with the current filters and patches. Nevertheless, extension to larger images is possible. Scalability of the current implementation has been presented and shows that larger images can be processed on the current FPGA with minor adjustments, but will ultimately be limited by the amount of internal memory available for buffering images and storing intermediate results.  To overcome this one could use a larger FPGA, use multiple FPGA's operating in parallel, reduce the number of S2 patches to free up memory, or change the model to use sparse features. 

To provide a fair comparison with the original HMAX model we used the same classifiers (boosting for binary and linear one-vs-all SVM for multi-class). Linear SVM classifiers remain the top choice for most HMAX implementations due to their computational simplicity and speed. {The choice of linear SVM classifiers is also supported by other work on discriminating between visual objects based on fMRI recordings of early stages of visual cortex \cite{FMRIclassifiers, FMRIclassifiersSupport1, FMRIclassifiersSupport2}.}   In our implementation we were able to run the classifier in the loop on a host PC without affecting the system throughput because classification was performed in parallel with feature extraction for the next image. Nevertheless, various classifiers can and have been implemented in FPGA \cite{classGMM, classNaiveBayes, classNN1, classNN2, classKNN}, including SVM \cite{classSVM}, and even a core generator for parameterized generation of your own classifier in FPGA \cite{classCORE}.

Comparison with other approaches shows that this is currently the fastest complete HMAX implementation and outperforms reported CNS \cite{GPUhmaxCNS} and custom \cite{CudaTimes2010} GPU implementations, as well as many FPGA implementations, although direct comparison with other FPGA implementations is not always possible. As more powerful GPU platforms become available these GPU implementations will achieve even better results, however the same can be said for FPGAs. The platform we have used (Xilinx Virtex 6 XC6VLX240T) is only in the middle of the range of the Virtex 6 family and is an entire technology generation behind the currently available Virtex 7 family.


\section{Conclusion}
We have shown how a neuromorphic bio-inspired hierarchical model of object recognition can be adapted for high speed implementation on a mid-range COTS FPGA platform. This implementation has a throughput of 190 images per second which is the fastest reported for a complete HMAX model. We have performed binary classification tests on popular Caltech 101 categories as well as on a more difficult Flickr dataset to show that adaption for FPGA does not have a significant effect on recognition accuracy. We have also shown that accuracy is not compromised on a multi-class classification task using Caltech 101.

\section*{Acknowledgment}
This work was partially supported by the Defense Advanced Research Projects Agency NeoVision2 program (government contract no. HR0011-10-C-0033) and the Research Program in Applied Neuroscience.



\bibliographystyle{IEEEtran}
%
\bibliography{TNNLS-2012-P-0824}
\begin{IEEEbiography}[{\includegraphics[width=1in,height=1.25in,clip,keepaspectratio]{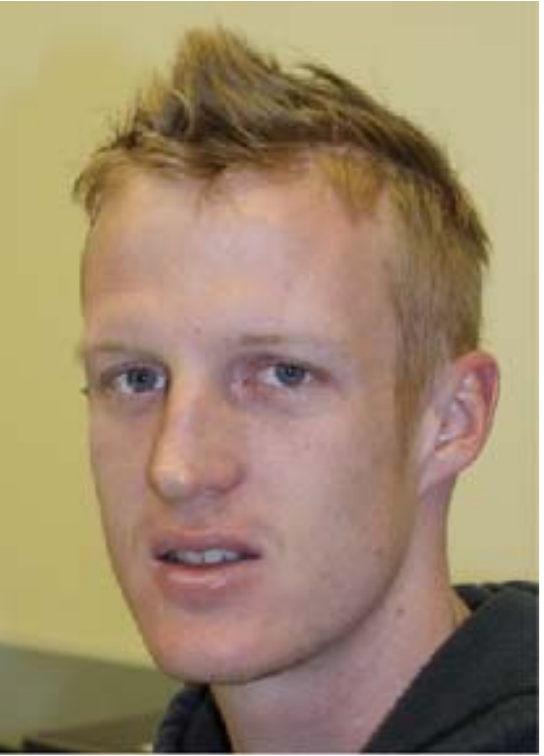}}]{Garrick Orchard}\input{Biographies/Garrick_Orchard.txt}
\end{IEEEbiography}

\begin{IEEEbiography}[{\includegraphics[width=1in,height=1.25in,clip,keepaspectratio]{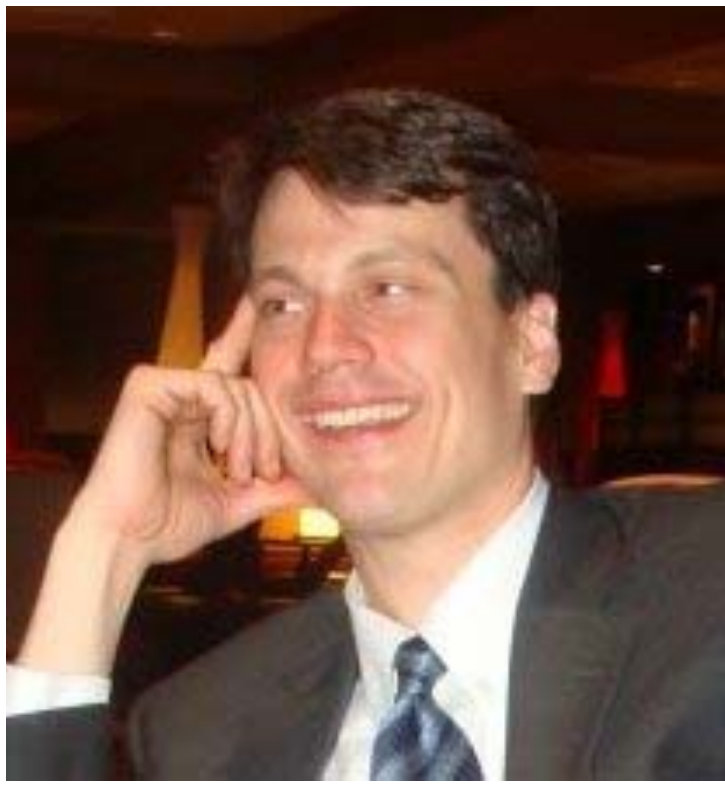}}]{Jacob G. Martin}\input{Biographies/Jacob_G_Martin.txt}
\end{IEEEbiography}

\begin{IEEEbiography}[{\includegraphics[width=1in,height=1.25in,clip,keepaspectratio]{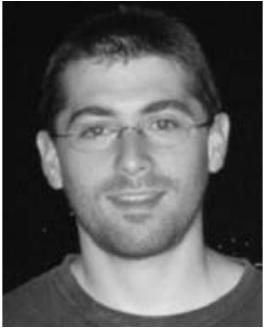}}]{R. Jacob Vogelstein}\input{Biographies/R_Jacob_Vogelstein.txt}
\end{IEEEbiography}

\begin{IEEEbiography}[{\includegraphics[width=1in,height=1.25in,clip,keepaspectratio]{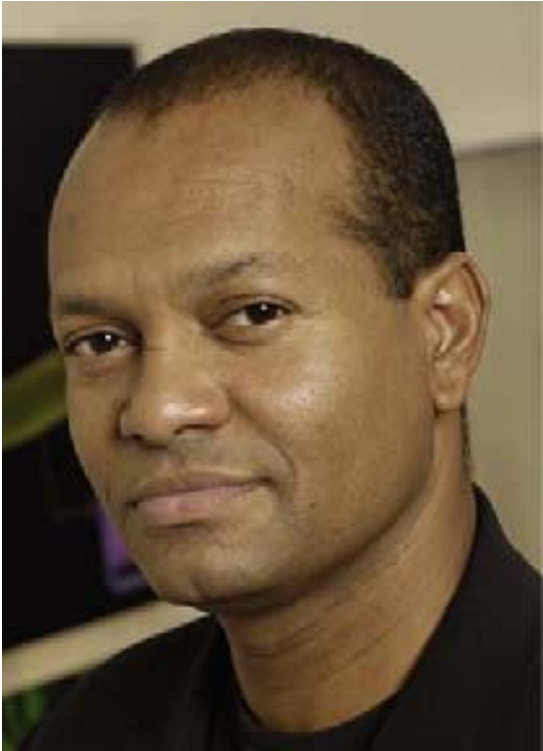}}]{Ralph Etienne-Cummings}\input{Biographies/Ralph_Etienne-Cummings.txt}
\end{IEEEbiography}

\end{document}

%% file: Biographies/Garrick_Orchard.txt
received the B.Sc. degree in electrical engineering from the University of Cape Town, South Africa in 2006 and the M.S.E. and Ph.D. degrees in electrical and computer engineering from Johns Hopkins University, Baltimore in 2009 and 2012 respectively. He was named a Paul V. Renoff fellow in 2007 and a Virginia and Edward M. Wysocki, Sr. fellow in 2011. He is also a recipient of the JHUAPL Hart Prize for Best Research and Development Project, and won the best live demonstration prize at the IEEE Biocas 2012 conference. He is currently a Postdoctoral Research Fellow at the Singapore Institute for Neurotechnology (SINAPSE) at the National University of Singapore where his research focuses on developing neuromorphic vision sensors and algorithms for real-time sensing on aerial platforms. His other research interests include mixed-signal very large scale integration (VLSI) design, compressive sensing, spiking neural networks, visual motion perception, and legged locomotion.

%% file: Biographies/Jacob_G_Martin.txt
received a Bachelor of Science and the PhD degree in Computer Science from the University of Georgia in 1999 and 2005. His first Postdoctoral Research Fellowship was at Trinity College Dublin in Ireland, where researched multisensory processing in collaboration with experimental neuroscientists.  His next position was as a Postdoctoral Fellow in the Department of Neuroscience at Georgetown University Medical Center in Washington, DC, where he worked with human psychophysics, EEG, and computational models of vision to explore the dynamics of visual processing in the human brain.  He is currently a Senior Staff Scientist in Applied Neuroscience at The Johns Hopkins University Applied Physics Laboratory.  His research is focused on cognitive neuroscience, human vision, brain computer interfaces, hybrid brain-machine vision systems, biologically-inspired machine vision, numerical analysis, spectral graph theory, information retrieval, and pattern recognition.

%% file: Biographies/R_Jacob_Vogelstein.txt
received the Sc.B. degree in neuroengineering from Brown University, Providence, RI, and the Ph.D. degree in biomedical engineering from the Johns Hopkins University School of Medicine, Baltimore, MD. He is the Assistant Program Manager for Applied Neuroscience at the Johns Hopkins University (JHU) Applied Physics Laboratory and an Assistant Research Professor in Electrical Engineering at the JHU's Whiting School. He has worked on neuroscience technology for over a decade, focusing primarily on neuromorphic systems and closed-loop brain–machine interfaces. His research has been featured in a number of prominent scientific and engineering journals including the IEEE Transactions on Neural Systems and Rehabilitation Engineering, the IEEE Transactions on Biomedical Circuits and Systems, and the IEEE Transactions on Neural Networks.

%% file: Biographies/Ralph_Etienne-Cummings.txt
received his B. Sc. in physics, 1988, from Lincoln University, Pennsylvania. He completed his M.S.E.E. and Ph.D. in electrical engineering at the University of Pennsylvania in 1991 and 1994, respectively. He is currently a professor of electrical and computer engineering, and computer science at Johns Hopkins University (JHU). He is the former Director of Computer Engineering at JHU and the Institute of Neuromorphic Engineering. He is also the Associate Director for Education and Outreach of the National Science Foundation (NSF) sponsored Engineering Research Centers on Computer Integrated Surgical Systems and Technology at JHU. He has served as Chairman of the IEEE Circuits and Systems (CAS) Technical Committee on Sensory Systems and on Neural Systems and Application. He was also the General Chair of the IEEE BioCAS 2008 Conference. He was also a member of Imagers, MEMS, Medical and Displays Technical Committee of the ISSCC Conference from 1999 – 2006. He is the recipient of the NSF's Career and Office of Naval Research Young Investigator Program Awards. In 2006, he was named a Visiting African Fellow and a Fulbright Fellowship Grantee for his sabbatical at University of Cape Town, South Africa. He was invited to be a lecturer at the National Academies of Science Kavli Frontiers Program, in 2007. He has won publication awards including the 2003 Best Paper Award of the EURASIP Journal of Applied Signal Processing and "Best Ph.D. in a Nutshell" at the IEEE BioCAS 2008 Conference, and has been recognized for his activities in promoting the participation of women and minorities in science, technology, engineering and mathematics. His research interest includes mixed signal VLSI systems, computational sensors, computer vision, neuromorphic engineering, smart structures, mobile robotics, legged locomotion and neuroprosthetic devices.

%% file: TNNLS-2012-P-0824.bbl
\begin{thebibliography}{10}
\providecommand{\url}[1]{#1}
\csname url@samestyle\endcsname
\providecommand{\newblock}{\relax}
\providecommand{\bibinfo}[2]{#2}
\providecommand{\BIBentrySTDinterwordspacing}{\spaceskip=0pt\relax}
\providecommand{\BIBentryALTinterwordstretchfactor}{4}
\providecommand{\BIBentryALTinterwordspacing}{\spaceskip=\fontdimen2\font plus
\BIBentryALTinterwordstretchfactor\fontdimen3\font minus
  \fontdimen4\font\relax}
\providecommand{\BIBforeignlanguage}[2]{{%
\expandafter\ifx\csname l@#1\endcsname\relax
\typeout{** WARNING: IEEEtran.bst: No hyphenation pattern has been}%
\typeout{** loaded for the language `#1'. Using the pattern for}%
\typeout{** the default language instead.}%
\else
\language=\csname l@#1\endcsname
\fi
#2}}
\providecommand{\BIBdecl}{\relax}
\BIBdecl

\bibitem{SIFT}
D.~Lowe, ``Object recognition from local scale-invariant features,''
  \emph{Proc. 7th IEEE Int. Conf. Computer Vision}, vol.~2, pp. 1150--1157,
  1999.

\bibitem{SURF}
H.~Bay, T.~Tuytelaars, and L.~{Van Gool}, ``{SURF}: Speeded-up robust
  features,'' \emph{9th European Conf. Computer Vision}, vol. 110, no.~3, pp.
  346--359, 2008.

\bibitem{MonkeyViewPaperclip}
N.~Logothetis, J.~Pauls, and T.~Poggio, ``Shape representation in the inferior
  temporal cortex of monkeys,'' \emph{Current Biology}, vol.~5, no.~5, pp.
  552--563, 1995.

\bibitem{HMAXOriginal}
M.~Riesenhuber and T.~Poggio, ``Are cortical models really bound by the
  �binding problem,'' \emph{Neuron}, vol.~24, pp. 87--93, 1999.

\bibitem{BestArchitectureForRecognition}
K.~Jarrett, K.~Kavukcuoglu, M.~Ranzato, and Y.~LeCun, ``What is the best
  multi-stage architecture for object recognition?'' \emph{Proc. 12th IEEE Int.
  Conf. Computer Vision}, pp. 2146--2153, Oct 2009.

\bibitem{HMAX2007}
T.~Serre, L.~Wolf, S.~Bileschi, M.~Riesenhuber, and T.~Poggio, ``Robust object
  recognition with cortex-like mechanisms,'' \emph{IEEE Trans. Pattern Analysis
  and Machine Intelligence}, vol.~29, no.~3, pp. 411--426, Mar 2007.

\bibitem{FeedforwardRapidCategorization}
T.~Serre, A.~Oliva, and T.~Poggio, ``A feedforward architecture accounts for
  rapid categorization,'' \emph{Proc. National Academy of Sciences}, vol. 104,
  no.~15, pp. 6424--6429, 2007.

\bibitem{AnimalNoAnimal}
S.~Thorpe, D.~Fize, and C.~Marlot, ``{Speed of processing in the human visual
  system},'' \emph{Nature}, vol. 381, no. 6582, pp. 520--522, Jun 1996.

\bibitem{MUTCH06}
J.~M. and D.~G.L., ``Multiclass object recognition with sparse, localized
  features,'' \emph{IEEE Conf. Computer Vision and Pattern Recognition}, pp.
  11--18, Jun 2006.

\bibitem{CUDA2008}
S.~Chikkerur, ``{CUDA} implementation of a biologically inspired object
  recognition system.'' \emph{{MIT} online}, 2008.

\bibitem{GPUhmaxCNS}
J.~Mutch, U.~Knoblich, and T.~Poggio, ``{CNS}: a {GPU}-based framework for
  simulating cortically-organized networks,'' Massachusetts Institute of
  Technology, Cambridge, MA, Tech. Rep., Feb 2010.

\bibitem{CudaTimes2010}
H.~Sedding, F.~Deger, H.~Dammertz, J.~Bouecke, and H.~Lensch, ``Massively
  parallel multiclass object recognition,'' \emph{Proc. Vision, Modeling, and
  Visualization Workshop}, pp. 251--257, 2010.

\bibitem{GPUobjectRecognition}
J.~Kim, E.~Park, X.~Cui, H.~Kim, and W.~Gruver, ``A fast feature extraction in
  object recognition using parallel processing on {CPU} and {GPU},'' \emph{IEEE
  Int. Conf. Systems, Man and Cybernetics}, pp. 3842--3847, Oct 2009.

\bibitem{GPUsift}
S.~Warn, W.~Emeneker, J.~Cothren, and A.~Apon, ``Accelerating {SIFT} on
  parallel architectures,'' \emph{IEEE Int. Conf. Cluster Computing}, pp. 1--4,
  Sept 2009.

\bibitem{GPUcudaRecognition}
R.~Uetz and S.~Behnke, ``Large-scale object recognition with {CUDA}-accelerated
  hierarchical neural networks,'' \emph{IEEE Int. Conf. Intelligent Computing
  and Intelligent Systems}, vol.~1, pp. 536--541, Nov 2009.

\bibitem{GPUrealTimerecognition}
M.~Ebner, ``A real-time evolutionary object recognition system,'' in
  \emph{Genetic Programming}, ser. Lecture Notes in Computer Science.\hskip 1em
  plus 0.5em minus 0.4em\relax Springer Berlin / Heidelberg, 2009, vol. 5481,
  pp. 268--279.

\bibitem{ASICKaist}
J.~Kim, M.~Kim, S.~Lee, J.~Oh, K.~Kim, and H.~Yoo, ``A 201.4 {GOPS} 496 {mW}
  real-time multi-object recognition processor with bio-inspired neural
  perception engine,'' \emph{IEEE J. Solid-State Circuits}, vol.~45, no.~1, pp.
  32--45, Jan 2010.

\bibitem{Impulse}
\BIBentryALTinterwordspacing
{Impulse Accelerated Technology Inc. Website}. [Online]. Available:
  \url{www.impulseaccelerated.com}
\BIBentrySTDinterwordspacing

\bibitem{CtoFPGAsurvey}
B.~Holland, M.~Vacas, V.~Aggarwal, R.~DeVille, I.~Troxel, and A.~D. George,
  ``Survey of {C}-based application mapping tools for reconfigurable
  computing,'' \emph{8th Int. Conf. Military and Aerospace Programmable Logic
  Devices}, Sept 2005.

\bibitem{NeuFlowISCAS2010}
C.~Farabet, B.~Martini, P.~Akselrod, S.~Talay, Y.~LeCun, and E.~Culurciello,
  ``Hardware accelerated convolutional neural networks for synthetic vision
  systems,'' \emph{Proc. IEEE Int. Symp. Circuits and Systems}, pp. 257--260,
  Jun 2010.

\bibitem{Lua}
R.~Ierusalimschy, L.~de~Figueiredo, and W.~Celes, \emph{Lua 5.1 Reference
  Manual}.\hskip 1em plus 0.5em minus 0.4em\relax Rio de Janeiro, Brazil:
  Lua.org, 2006.

\bibitem{PENN1}
A.~Maashri, M.~DeBole, M.~Cotter, N.~Chandramoorthy, Y.~Xiao, V.~Narayanan, and
  C.~Chakrabarti, ``Accelerating neuromorphic vision algorithms for
  recognition,'' in \emph{49th IEEE Design Automation Conference}, Jun 2012,
  pp. 579--584.

\bibitem{PENN2}
S.~Kestur, M.~S. Park, J.~Sabarad, D.~Dantara, V.~Narayanan, Y.~Chen, and
  D.~Khosla, ``Emulating mammalian vision on reconfigurable hardware,'' in
  \emph{20th IEEE Annu. Int. Symp. Field-Programmable Custom Computing
  Machines}, May 2012, pp. 141--148.

\bibitem{PENN3}
M.~DeBole, Y.~Xiao, C.-L. Yu, A.~Maashri, M.~Cotter, C.~Chakrabarti, and
  V.~Narayanan, ``{FPGA}-accelerator system for computing biologically inspired
  feature extraction models,'' in \emph{45th Asilomar Conf. Signals, Systems
  and Computers}, Nov 2011, pp. 751--755.

\bibitem{PENN4}
A.~Al~Maashri, M.~DeBole, C.-L. Yu, V.~Narayanan, and C.~Chakrabarti, ``A
  hardware architecture for accelerating neuromorphic vision algorithms,'' in
  \emph{IEEE Workshop Signal Processing Systems}, Oct 2011, pp. 355--360.

\bibitem{PENN5}
M.~S. Park, S.~Kestur, J.~Sabarad, V.~Narayanan, and M.~Irwin, ``An
  {FPGA}-based accelerator for cortical object classification,'' in
  \emph{Design, Automation Test in Europe Conf.}, Mar 2012, pp. 691--696.

\bibitem{PENN6}
J.~Sabarad, S.~Kestur, M.~S. Park, D.~Dantara, V.~Narayanan, Y.~Chen, and
  D.~Khosla, ``A reconfigurable accelerator for neuromorphic object
  recognition,'' in \emph{17th Asia and South Pacific Design Automation Conf.},
  Feb 2012, pp. 813--818.

\bibitem{Lucas_Kanade_1981}
B.~Lucas and T.~Kanade, ``An iterative image registration technique with an
  application to stereo vision,'' \emph{Proc. DARPA Image Understanding
  Workshop}, pp. 121--130, Apr 1981.

\bibitem{FPGAopticalFlow}
Z.~Wei, D.~Lee, and B.~Nelson, ``{FPGA}-based real-time optical flow algorithm
  design and implementation,'' \emph{J. Multimedia}, vol.~2, no.~5, 2007.

\bibitem{SIFTFPGA1}
L.~Yao, H.~Feng, Y.~Zhu, Z.~Jiang, D.~Zhao, and W.~Feng, ``An architecture of
  optimised {SIFT} feature detection for an {FPGA} implementation of an image
  matcher,'' \emph{Int. Conf. Field-Programmable Technology}, pp. 30--37, Dec
  2009.

\bibitem{SIFTFPGA2}
V.~Bonato, E.~Marques, and G.~Constantinides, ``A parallel hardware
  architecture for scale and rotation invariant feature detection,'' \emph{IEEE
  Trans. Circuits and Systems for Video Technology}, vol.~18, no.~12, pp.
  1703--1712, Dec 2008.

\bibitem{SURFFPGA}
J.~Svab, T.~Krajnik, J.~Faigl, and L.~Preucil, ``{FPGA} based {S}peeded {U}p
  {R}obust {F}eatures,'' \emph{IEEE Int. Conf. Technologies for Practical Robot
  Applications}, pp. 35--41, Nov 2009.

\bibitem{SpatioTemporalTracking}
K.~Cannons and R.~Wildes, ``Spatiotemporal oriented energy features for visual
  tracking,'' in \emph{Computer Vision ACCV 2007}, ser. Lecture Notes in
  Computer Science, Y.~Yagi, S.~Kang, I.~Kweon, and H.~Zha, Eds.\hskip 1em plus
  0.5em minus 0.4em\relax Springer Berlin / Heidelberg, 2007, vol. 4843, pp.
  532--543.

\bibitem{SpatioTemporalFPGASegmentation}
K.~Ratnayake and A.~Amer, ``An {FPGA}-based implementation of spatio-temporal
  object segmentation,'' \emph{IEEE Int. Conf. Image Processing}, pp.
  3265--3268, Oct 2006.

\bibitem{DepthPerceptionVergenceBert}
E.~Tsang, S.~Lam, Y.~Meng, and B.~Shi, ``Neuromorphic implementation of active
  gaze and vergence control,'' \emph{Proc. IEEE Int. Symp. Circuits and
  Systems}, pp. 1076--1079, May 2008.

\bibitem{FixedVSfloating}
A.~Savich, M.~Moussa, and S.~Areibi, ``The impact of arithmetic representation
  on implementing {MLP-BP} on {FPGAs}: A study,'' \emph{IEEE Trans. Neural
  Networks}, vol.~18, no.~1, pp. 240--252, Jan 2007.

\bibitem{Boltzmann}
D.~Le~Ly and P.~Chow, ``High-performance reconfigurable hardware architecture
  for restricted boltzmann machines,'' \emph{IEEE Trans. Neural Networks},
  vol.~21, no.~11, pp. 1780--1792, Nov 2010.

\bibitem{SpikingFPGA}
M.~Pearson, A.~Pipe, B.~Mitchinson, K.~Gurney, C.~Melhuish, I.~Gilhespy, and
  M.~Nibouche, ``Implementing spiking neural networks for real-time
  signal-processing and control applications: A model-validated {FPGA}
  approach,'' \emph{IEEE Trans. Neural Networks}, vol.~18, no.~5, pp.
  1472--1487, Sept 2007.

\bibitem{MultiLayerMultiPlex}
S.~Himavathi, D.~Anitha, and A.~Muthuramalingam, ``Feedforward neural network
  implementation in {FPGA} using layer multiplexing for effective resource
  utilization,'' \emph{IEEE Trans. Neural Networks}, vol.~18, no.~3, pp.
  880--888, May 2007.

\bibitem{classGMM}
M.~Shi, A.~Bermak, S.~Chandrasekaran, and A.~Amira, ``An efficient {FPGA}
  implementation of gaussian mixture models-based classifier using distributed
  arithmetic,'' in \emph{13th IEEE Int. Conf. Electronics, Circuits and
  Systems}, Dec 2006, pp. 1276--1279.

\bibitem{classNN1}
F.~Benrekia, M.~Attari, A.~Bermak, and K.~Belhout, ``{FPGA} implementation of a
  neural network classifier for gas sensor array applications,'' in \emph{6th
  Int. Multi-Conf. Systems, Signals and Devices}, Mar 2009, pp. 1--6.

\bibitem{classNN2}
T.~Nguyen, K.~Chandan, B.~Ahmad, and K.~Yap, ``{FPGA} implementation of neural
  network classifier for partial discharge time resolved data from magnetic
  probe,'' in \emph{Int. Conf. Advanced Power System Automation and
  Protection}, vol.~1, Oct 2011, pp. 451--455.

\bibitem{classNaiveBayes}
H.~Meng, K.~Appiah, A.~Hunter, and P.~Dickinson, ``{FPGA} implementation of
  naive bayes classifier for visual object recognition,'' in \emph{IEEE Conf.
  Computer Vision and Pattern Recognition Workshops}, Jun 2011, pp. 123--128.

\bibitem{classKNN}
H.~M. Hussain, K.~Benkrid, and H.~Seker, ``An adaptive implementation of a
  dynamically reconfigurable k-nearest neighbour classifier on {FPGA},'' in
  \emph{NASA/ESA Conf. Adaptive Hardware and Systems}, Jun 2012, pp. 205--212.

\bibitem{classSVM}
A.~Fazakas, M.~Neag, and L.~Festila, ``Block {RAM} versus distributed {RAM}
  implementation of {SVM} classifier on {FPGA},'' in \emph{IEEE Conf. Applied
  Electronics}, Sept 2006, pp. 43--46.

\bibitem{classCORE}
D.~Anguita, L.~Carlino, A.~Ghio, and S.~Ridella, ``A {FPGA} core generator for
  embedded classification systems,'' in \emph{J. Circuits, Systems and
  Computers}, vol.~20, no.~2, Apr 2011, pp. 263--282.

\bibitem{FMRIclassifiers}
M.~Misaki, Y.~Kim, P.~Bandettini, and N.~Kriegeskorte, ``Comparison of
  multivariate classifiers and response normalizations for pattern-information
  {fMRI},'' \emph{NeuroImage}, vol.~53, no.~1, pp. 103--118, Oct 2010.

\bibitem{FMRIclassifiersSupport1}
D.~Cox, R.~Savoy \emph{et~al.}, ``Functional magnetic resonance imaging
  ({fMRI}) brain reading: detecting and classifying distributed patterns of
  {fMRI} activity in human visual cortex,'' \emph{NeuroImage}, vol.~19, no.~2,
  pp. 261--270, Jun 2003.

\bibitem{FMRIclassifiersSupport2}
S.~LaConte, S.~Strother, V.~Cherkassky, J.~Anderson, and X.~Hu, ``Support
  vector machines for temporal classification of block design {fMRI} data,''
  \emph{NeuroImage}, vol.~26, no.~2, pp. 317 -- 329, 2005.

\bibitem{Caltech101}
L.~Fei-Fei, R.~Fergus, and P.~Perona, ``Learning generative visual models from
  few training examples: An incremental {B}ayesian approach tested on 101
  object categories,'' \emph{Computer Vision and Pattern Recognition Workshop},
  p. 178, Jun 2004.

\bibitem{HUBELWeisel}
D.~Hubel and T.~Weisel, ``Receptive fields, binocular interaction, and function
  architecture in the cat's visual cortex,'' \emph{J Physiol (Lond.)}, vol.
  160, pp. 106--154, 1962.

\bibitem{antoniou2000digital}
A.~Antoniou, \emph{Digital Filters: Analysis, Design, and Applications}.\hskip
  1em plus 0.5em minus 0.4em\relax McGraw-Hill, 2000.

\bibitem{GentleBoost}
J.~Friedman, T.~Hastie, and R.~Tibshirani, ``{Special invited paper. Additive
  logistic regression: A statistical view of boosting},'' \emph{Ann.
  Statistics}, vol.~28, 2000.

\bibitem{SVMmulticlassTraining}
T.~Joachims, T.~Finley, and C.-N.~J. Yu, ``Cutting-plane training of structural
  {SVMs},'' \emph{J. Mach. Learn. Res.}, vol.~77, no.~1, pp. 27--59, Oct 2009.

\bibitem{SVMmulticlass}
K.~Crammer and Y.~Singer, ``On the algorithmic implementation of multiclass
  kernel-based vector machines,'' \emph{J. Mach. Learn. Res.}, vol.~2, pp.
  265--292, Mar 2002.

\bibitem{HMAXwebsite}
\BIBentryALTinterwordspacing
{HMAX Website}. [Online]. Available:
  \url{http://riesenhuberlab.neuro.georgetown.edu/hmax.html}
\BIBentrySTDinterwordspacing

\bibitem{HMAXsparse}
J.~Mutch and D.~Lowe, ``Object class recognition and localization using sparse
  features with limited receptive fields,'' \emph{International Journal of
  Computer Vision}, vol.~80, pp. 45--57, Oct 2008.

\end{thebibliography}
